\documentclass[letterpaper, 10 pt, conference]{ieeeconf}  % Comment this line out if you need a4paper

\IEEEoverridecommandlockouts                              % This command is only needed if 
\usepackage{graphics} % for pdf, bitmapped graphics files
\usepackage{amsmath, amsfonts} % assumes amsmath package installed
\usepackage{cite}
\usepackage{booktabs}
\usepackage{multirow}
\usepackage{tabularx}
\usepackage[nolist,nohyperlinks]{acronym}
\usepackage{cleveref}
\usepackage{graphicx}
\usepackage{xcolor}
\usepackage{siunitx}

\acrodef{rl}[RL]{Reinforcement Learning}
\acrodef{appo}[APPO]{Asynchronous Proximal Policy Optimization}
\acrodef{gru}[GRU]{Gated Recurrent Unit}
\acrodef{mlp}[MLP]{Multi-Layer Perceptron}
\acrodef{dce}[DCE]{Deep Collision Encoder}
\acrodef{lnn}[LNN]{Liquid Neural Network}
\acrodef{wta}[WTA]{Winner-Takes-All}
\acrodef{ior}[IOR]{Inhibition of Return}
\acrodef{rnn}[RNN]{Recurrent Neural Network}
\acrodef{ltc}[LTC]{Liquid Time-Constant}
\acrodef{cfc}[CfC]{Closed-Form Continuous-time}
\acrodef{dtw}[DTW]{Dynamic Time Warping}
\acrodef{tde}[TDE]{Time Delay Embedding}

\title{\LARGE \bf
Fast Human Attention Prediction for Fixation-guided\\ Active Perception in Autonomous Navigation
}

\author{Fatma Youssef Mohammed$^{*}$, Grzegorz Malczyk$^{*}$, and Kostas Alexis
\thanks{$^*$ The authors contributed equally.}
\thanks{This work was supported by the Horizon Europe Grant No. 101120732 and the Research Council of Norway projects under Award NO-338694, NO-357451. The authors are with the Department of Engineering Cybernetics, Norwegian University of Science and Technology (NTNU), Norway.}
\thanks{\tt\small fatma.y.m.a.e.f.mohammed@ntnu.no}
}

\begin{document}

\maketitle
\thispagestyle{empty}
\pagestyle{empty}

%%%%%%%%%%%%%%%%%%%%%%%%%%%%%%%%%%%%%%%%%%%%%%%%%%%%%%%%%%%%%%%%%%%%%%%%%%%%%%%%
\begin{abstract}
% While human visual attention efficiently processes scenes via structured, feature-driven scanpaths, replicating this active perception within resource-constrained robot autonomy remains a significant challenge. 
Human visual attention relies on structured scanpaths to efficiently process scenes, yet instilling this behavior into robot autonomy is in its infancy and hindered by the high computational costs of existing predictive models.
To address this, we introduce GazeLNN, a computationally lightweight scanpath prediction model that leverages Liquid Neural Networks as its recurrent engine and employs MobileNetV3 for feature extraction. Operating auto-regressively, the architecture predicts sequential fixation heatmaps conditioned on the current visual stimulus and fixation history. Despite requiring only 0.61 GFLOPs, GazeLNN achieves state-of-the-art performance on the MIT Low Resolution dataset achieving 0.47 ScanMatch score. It outperforms existing recurrent baselines across diverse evaluation metrics, while reducing computational costs by 99.40\% and accelerating inference by up to six times. To investigate the role of human attention modeling in robot autonomy and demonstrate the practical utility of this highly efficient architecture, we integrate GazeLNN into an active camera-robot control policy trained via Reinforcement Learning. This integration enables human-fixation-guided perception during autonomous navigation, validated through successful real-world deployments on an aerial robot.

% Human visual attention follows structured scanpaths driven by low-level visual features, yet efficiently modeling this behavior and instilling it within robot autonomy remains a challenge. Responding to this fact, we first present a computationally lightweight scanpath prediction model that leverages \acp{lnn}, as the recurrent component and MobileNetV3 as the feature extractor backbone. The proposed model adopts an auto-regressive formulation, predicting sequential fixation heatmaps conditioned on the input image and fixation history. The proposed model requires only $0.61$ GFLOPs and $15.24$ million parameters, achieving an inference speed of $6.84~\text{ms}$ per frame. The proposed model achieves state-of-the-art performance on the MIT-Low Resolution dataset, outperforming existing recurrent-based methods across the majority of evaluation metrics while reducing computational cost by up to $99.40\%$ and improving inference speed by up to $6.42\times$ compared to the second-best baseline. Targeting the uptake of such models in robot autonomy, we integrate our model into an active camera control policy via \ac{rl}, enabling human-like gaze-guided perception in autonomous robotic navigation. We validate the full pipeline through successful real-world deployments on an aerial robot. 
%This represents the first use of human attention modeling in active robotic perception.
\end{abstract}

%%%%%%%%%%%%%%%%%%%%%%%%%%%%%%%%%%%%%%%%%%%%%%%%%%%%%%%%%%%%%%%%%%%%%%%%%%%%%%%%
\section{INTRODUCTION}
%update numbers based on new results
In autonomous robotics, processing high-resolution visual data across the entire field of view is often computationally prohibitive, especially for agile platforms with strict onboard resource constraints. To overcome this sensory bottleneck, roboticists frequently draw inspiration from the human visual system, which employs active perception to selectively allocate processing resources. Unlike standard robot vision, the human eye employs actuation and relies on high-acuity foveal vision for detailed inspection combined with lower-resolution peripheral vision for contextual awareness~\cite{stewart2020review}. By executing rapid gaze shifts - known as ``saccades'' -interspersed with brief pauses, or fixations, humans efficiently sample salient regions of a scene. This sequence of fixations, termed a fixation scanpath, is largely driven by bottom-up visual features such as contrast, color, and orientation~\cite{itti2002model,koch1987shifts} and thus represents a selective information-guided behavior. Emulating this stimulus-driven attention mechanism in robots can enable a more efficient, task-relevant allocation of sensory and computational bandwidth alongside driving active vision behaviors.

\begin{figure}[t]
    \centering
    \includegraphics[width=0.97\columnwidth]{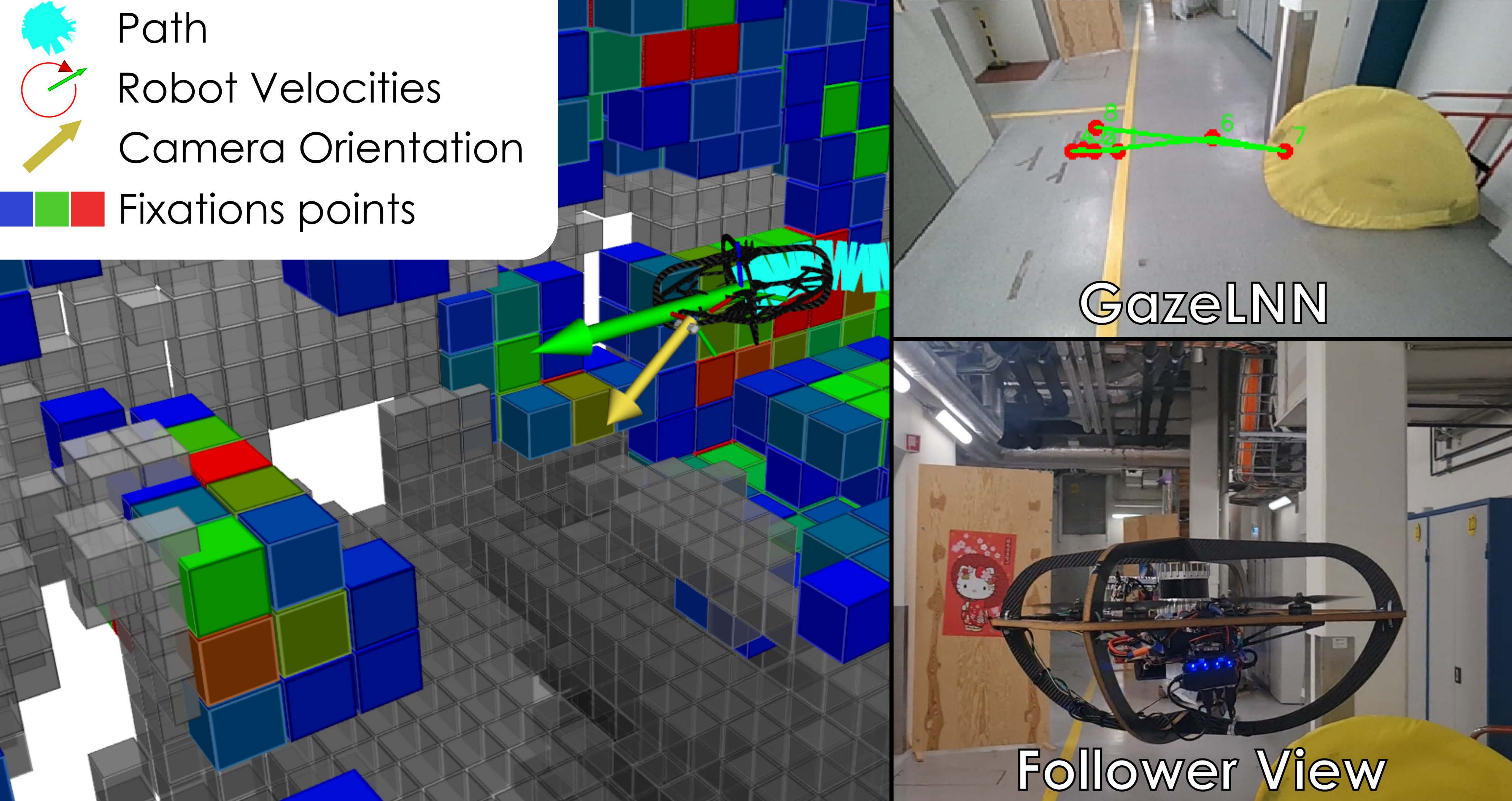}
    \vspace{-2ex}
    \caption{Real-world deployment of our fixation-guided active perception system. The aerial platform performs autonomous navigation based on an active camera-robot policy informed by the GazeLNN visual attention model.}
    \label{fig:title_image}
    \vspace{-4ex}
\end{figure}

While recent advances in deep learning have significantly improved the accuracy of scanpath prediction, they predominantly rely on heavy architectures. The computational demands and slow inference speeds of modern Transformer and deep recurrent models render them ill-suited for real-time, embodied robotic deployment. To address this gap, we propose modeling human visual attention using a lightweight architecture powered by \acp{lnn}~\cite{hasani2021liquid}. As brain-inspired, continuous-time models, \acp{lnn} provide a highly adaptive alternative to conventional \acp{rnn}. Their input-dependent temporal dynamics allow them to flexibly capture the stochastic nature of human gaze behavior. Furthermore, \acp{lnn} have demonstrated exceptional expressive power, achieving faster inference and significantly reduced computational overhead in time-series and control tasks~\cite{hasani2021liquid,hasani2022closed}. 
%To our knowledge, this work represents the first application of an \ac{lnn} for real-time scanpath prediction.
Our proposed model, GazeLNN, achieves state-of-the-art predictive performance while operating at a fraction of the computational cost of existing baselines. Requiring only $0.61$ GFLOPs and $15.24$ million parameters, GazeLNN achieves a real-time inference speed of $6.84$~ms per frame  for predicting an $8$-fixation scanpath on an NVIDIA RTX 3500 Ada GPU. Evaluated on the MIT Low Resolution dataset, GazeLNN outperforms all recurrent-based baselines across all reported metrics, improving the ScanMatch score by $34.29\%$, Levenshtein Distance by $2.18\%$, Hausdorff Distance by $5.09\%$, Fréchet Distance by $7.92\%$, FastDTW by $11.85\%$, and Time Delay Embedding by $37.81\%$. Additionally, GazeLNN reduces computational costs by $99.40\%$ and runs $6.42\times$ faster than the second-best baseline, making it exceptionally well-suited for resource-constrained robotics.

To demonstrate its practical utility, we integrate GazeLNN into an \ac{rl}-based active camera-robot control policy. While recent active vision methods (e.g.,~\cite{malczyk2026reinforcement}) lack inherent attention modeling, our approach explicitly optimizes for a human-fixation-driven behavior by directing the camera toward predicted salient regions. Real-world experiments employing an aerial robot with an actuated camera confirm this significantly enhances spatial awareness: compared to a static (forward-facing) camera baseline, the actuated camera policy accumulates nearly $50\%$ more voxels in its global map and yields an eight-fold increase in observing salient-relevant regions, ultimately building a richer, more robust scene representation for autonomous navigation.

The remainder of this paper is structured as follows: \Cref{sec:related work} reviews related work, while \Cref{sec:methods} details the proposed GazeLNN. \Cref{sec:active_perception_rl} introduces the active vision-based navigation \ac{rl}. \Cref{sec: results} analyzes the network's performance, including comprehensive ablation studies. \Cref{sec:experiments} presents the experimental evaluation of the \ac{rl}. Finally, conclusions are drawn in \Cref{sec: conclusion}.

\section{RELATED WORK}
\label{sec:related work}

Our contribution relates to the domains of human attention modeling and its integration into robot autonomy systems. 

\subsection{Human Attention Models}
Early attention models focused on static saliency maps using hand-crafted features~\cite{itti2002model,koch1987shifts} or deep learning~\cite{linardos2021deepgaze,hosseini2025sum}. Since these architectures inherently fail to capture the actual underlying sequential dynamics of human viewing behavior, scanpath prediction models were introduced. Foundational approaches derived fixation sequences using \ac{wta} selection and \ac{ior}~\cite{itti2002model}, while modern deep learning models condition the next fixation on both the image stimulus and historical gaze data. For instance~\cite{kummerer2022deepgaze} incorporates history spatially without explicitly modeling temporal recurrence. Simultaneously, \ac{rnn}-based architectures, such as auto-regressive ConvLSTMs, are widely utilized to explicitly capture both spatial and temporal dependencies within the scanpath~\cite{chen2018scanpath,chen2021predicting,martin2024tspm}. 

More recently, Transformers have established new predictive benchmarks~\cite{yang2024unifying,mondal2023gazeformer}. However, their profound computational demands make them impractical for resource-constrained robotic platforms. Interestingly, a recent study~\cite{mohammed2025unified} showed that freezing the most compute-intensive components of an attention Transformer~\cite{yang2024unifying} yields performance comparable to fully end-to-end trained models. This crucial finding suggests that over-parameterized networks are unnecessary for this task, directly motivating our proposed parsimonious and computationally efficient \ac{lnn}-based architecture.

\subsection{Attention Models in Robot Autonomy}

A niche set of works have investigated the problem of fusing human attention models in robot autonomy. 
%Early efforts, such as~\cite{frintrop2008attentional}, exploited saliency models to support visual Simultaneous Localization And Mapping (SLAM), while~\cite{chang2010mobile} presented an integrated vision-based navigation and localization system combining gist-based scene classification with saliency-driven landmark detection. In the context of safety-critical autonomy,~\cite{guo2021motion} proposed a motion-saliency-driven collision avoidance solution computing position- and direction-based object saliency to selectively identify dangerous obstacles and predict their future states. Focusing on active visual search,~\cite{rasouli2020attention} fuses  bottom-up and top-down saliency into a belief model in order to reduce search time in cluttered environments, and~\cite{dang2018visual} integrates a visual saliency model in robot exploration in order to acquire high-quality views in regions of interest. On the survey side,~\cite{potapova2017survey} analyzes three decades of research on 3D visual attention and outlines how such models can be integrated within detection, navigation, and active perception~\cite{bajcsy2018revisiting} loops in robotics. More recently,~\cite{liang2024visarl} leverages a small set of human-annotated saliency maps to train a predictor and pretrain visual encoders with RGB-saliency inputs, yielding saliency-conditioned latent representations improving learning for manipulation tasks.
Early efforts leveraged saliency to support visual Simultaneous Localization and Mapping (SLAM)~\cite{frintrop2008attentional} and landmark detection~\cite{chang2010mobile}. In safety-critical contexts, motion saliency has been used to selectively identify dangerous obstacles and predict their trajectories~\cite{guo2021motion}. For active visual search and exploration, fusing bottom-up and top-down saliency can reduce search times in cluttered environments~\cite{rasouli2020attention} and help acquire high-quality views of interesting regions~\cite{dang2018visual}. Recently, saliency maps have been utilized to pretrain visual encoders, yielding conditioned latent representations that improve downstream learning for manipulation~\cite{liang2024visarl}. A survey on how 3D visual attention can be integrated across detection, navigation, and active perception~\cite{bajcsy2018revisiting} is found in~\cite{potapova2017survey}.

Most relevant to our research are efforts coupling attention with active vision and navigation~\cite{rasouli2017integrating,ma2018saliency,pfeiffer2022visual}. The work in~\cite{rasouli2017integrating} introduced a unified framework integrating Bayesian viewpoint selection, 3D bottom-up saliency, and object-based navigation, enabling a robot with a pan-tilt camera to minimize search time and travel distance. In reactive navigation,~\cite{ma2018saliency} mapped CNN-based saliency directly to coordinated obstacle avoidance maneuvers. Finally,~\cite{pfeiffer2022visual} demonstrated that a gaze prediction model trained on human eye-tracking data from drone racing can improve quadrotor control when its encoded features are integrated into an end-to-end policy.

% Most relevant to our work are the efforts in~\cite{rasouli2017integrating,ma2018saliency,pfeiffer2022visual}. The method in~\cite{rasouli2017integrating} proposes a unified active vision framework integrating (i) Bayesian viewpoint selection over a 3D occupancy map, (ii) bottom-up saliency projected into 3D space, and (iii) top-down object-bsaed navigation enabling a pan-tilt stereo-equipped robot to active control its gaze and navigation to reduce search actions, time, and travel distance. In~\cite{ma2018saliency}, a vision-based reactive obstacle navigation framework is proposed combining CCN-based saliency detection and a policy trained to learn a direct mapping from visually-inferred obstacle states to coordinated maneuvering. In~\cite{pfeiffer2022visual}, the authors introduced a ResNet-18-based gaze prediction model trained explicitly on human First-Person-View (FPV) drone racing eye-tracking data and integrated its encoder features in an end-to-end quadrotor controller thus demonstrating that human visual attention conditioning can improve racing performance. 

Compared to these works, our approach distinctly advances the domain in two key ways: (i) by introducing a highly efficient \ac{lnn}-based architecture that achieves state-of-the-art bottom-up visual attention prediction; and (ii) by integrating this lightweight model into an \ac{rl} framework that explicitly optimizes active camera control for dynamic, human-fixation-guided behavior during autonomous flight.

%is distinct in both that it proposes a novel \ac{lnn}-based model geared towards top performance in bottom-up visual attention and in its integration within \ac{rl} for robot navigation exploiting active camera control.

% \begin{figure*}[t]
%     \centering
%     \includegraphics[width=0.99\textwidth]{Figures/lnn_modified.png}
%     \vspace{-2ex}
%     \caption{\textbf{Overall of GazeLNN model}. The input image is first processed by a feature extraction backbone and concatenated with the previous fixation, represented as a Gaussian heatmap (with the initial fixation fixed at the image center). The fixation heatmap is augmented with coordinate convolution channels to enrich spatial representation. An initial hidden state (initialized to zeros) is concatenated with these inputs and passed to a CfC cell to predict the next fixation heatmap. The CfC output is projected to a single channel and upsampled to the input image resolution. The hidden state is updated at each step, and the predicted heatmap is fed back as input for the subsequent prediction}
%     \label{fig: LNN arch}
%     \vspace{-4ex}
% \end{figure*}

\begin{figure*}[t]
    \centering
    \includegraphics[width=0.99\textwidth]{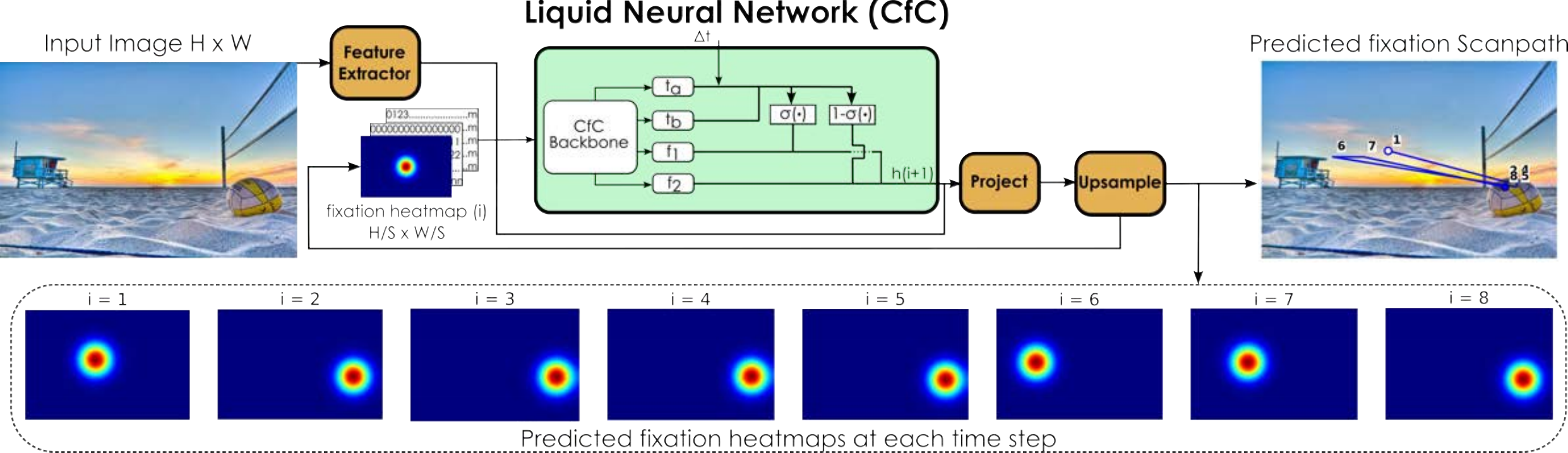}
    \vspace{-2ex}
    \caption{Overview of the GazeLNN model. The input image is first processed by a feature extraction backbone and concatenated with the previous fixation, represented as a Gaussian heatmap (with the initial fixation fixed at the image center). The fixation heatmap is augmented with the CoordConv layer~\cite{liu2018intriguing} to enrich spatial representation. The hidden state (initialized to zeros) is concatenated with these inputs and passed to a liquid neural network (CfC variant) to predict the next fixation heatmap. The CfC output is projected and fed back to CfC for the subsequent prediction.}
    \label{fig: LNN arch}
    \vspace{-4ex}
\end{figure*}

\section{SCANPATH PREDICTION APPROACH}
\label{sec:methods}
This section provides an overview of the network architecture used for scanpath prediction, the used dataset and training implementation details. 
\subsection{Network Architecture}
\label{sec: Network architecture}

% The overall model architecture is shown in~\Cref{fig: LNN arch}. We adopted an auto-regressive approach to predict fixation scanpaths, where the predicted fixation at step $i$ is used to infer fixation $i+1$. 

The overall architecture of the proposed model is illustrated in~\Cref{fig: LNN arch}. 
Our framework predicts fixation scanpaths in an auto-regressive manner, where the fixation predicted at step $i$ is used to infer the next fixation at step $i+1$. 
Given an input image, the model extracts visual features, integrates them with the current fixation heatmap, and iteratively predicts the next fixation using a Liquid Neural Network (LNN), as the recurrent module. This process allows the model to progressively generate a sequence of fixations forming a scanpath.

First, the input image is processed by a feature extraction backbone, specifically MobileNetV3, to extract the relevant features. MobileNetV3 is chosen due to its efficient architecture and strong performance in extracting visual features while maintaining low computational cost~\cite{howard2019searching}. The extracted features have a spatial resolution of $\frac{H}{S} \times \frac{W}{S}$, where $H$ and $W$ represent the image height and width, respectively, and $S$ represents the downsampling size. The choice of the feature extractor is analyzed in the ablation study (section~\ref{sec: backbone ablation}).

The extracted image features are concatenated with the fixation heatmap and provided as input to the recurrent module to predict the next fixation. The first fixation is fixed at the image center. Following prior work~\cite{chen2021predicting,martin2024tspm,chen2018scanpath}, fixations are represented as heatmaps instead of raw $(x,y)$ coordinates, as this representation provides a richer spatial structure for convolution models to learn. The fixation heatmap is combined with a CoordConv layer~\cite{liu2018intriguing}, which provides convolutional layers with explicit access to their input coordinates through additional $x$ and $y$ coordinate channels, improving spatial representation~\cite{liu2018intriguing,martin2024tspm}.

The concatenated image features and fixation representation are then combined with the hidden state (initialized to zeros), and passed to a \ac{lnn}, specifically the \ac{cfc} variant~\cite{hasani2022closed}. The choice of the \ac{cfc} is due to its faster training and inference time compared to the \ac{ltc} variant~\cite{hasani2021liquid}.

The combined input is first processed by a shared fully connected backbone consisting of $1024$ units followed by a LeCunTanh activation~\cite{lecun2002efficient}. The output is then passed to four parallel fully connected layers with $512$ units each, denoted as $f_1$, $f_2$, $t_a$, and $t_b$. The outputs of $f_1$ and $f_2$ are passed through a $\tanh$ activation, while $t_a$ and $t_b$ are used to compute a time-dependent gating signal. This gating signal is obtained by applying a sigmoid activation to a linear combination of $t_a$, $t_b$, and the elapsed time between fixations $\Delta t$. The resulting gate determines how the outputs of $f_1$ and $f_2$ are combined to update the hidden state of the CfC module. The hidden state update of the CfC module is defined as~\cite{hasani2022closed}

\begin{equation}
\begin{aligned}
h_{i+1} &= (1 - \sigma(t_a \mathbf{\Delta t} + t_b)) 
          \odot \tanh({f_1}(\mathbf{{x}_t})) \\
        &\quad + \sigma(t_a \mathbf{\Delta t} + t_b) 
          \odot \tanh({f_2}(\mathbf{{x}_t}))
\end{aligned}
\end{equation}
where $\sigma(\cdot)$ denotes the sigmoid activation, and $\Delta t$ represents the elapsed time between fixations obtained from the ground-truth fixation durations.

At each time step, the new hidden state of the CfC module is projected to the downsampled-size fixation map $\frac{H}{S} \times \frac{W}{S}$ and augmented with a CoordConv layer~\cite{liu2018intriguing} then fed back to \ac{cfc} module to predict the subsequent fixation heatmap. The predicted fixation map is further upsampled to the input image size ${H} \times{W}$ and divided by the sum of its values, yielding the next fixation heatmap distribution. The location of the maximum value of the predicted heatmap represents the next fixation and is used to compose the fixation scanpath.

% At each time step, the Conv-CfC cell output is fed to the projection layer, $\operatorname{Conv}_{\text{project}}$, that applies a convolution kernel of $1\times1$  and outputs a single channel. Then the output is upsampled to the input image size and divided by the sum of its values, yielding the next fixation heatmap distribution. The location of the maximum value of the predicted heatmap, represents the next fixation.

% At each time step, the CfC output is fed to a fully connected layer that projects the output back to $\frac{H}{S} \times \frac{W}{S}$. Then the output is upsampled to the input image size and divided by the sum of its values, yielding the next fixation heatmap distribution. The location of the maximum value of the predicted heatmap, represents the next fixation.

\subsection{Implementation Details}
\label{sec: Implementation Details}
Following~\cite{sun2019visual,martin2024tspm}, the model is trained on the OSIE dataset~\cite{xu2014predicting}. The OSIE dataset~\cite{xu2014predicting} comprises $700$ images, each viewed by $15$ subjects. The dataset is divided into $80\%$ training, $10\%$ validation, and $10\%$ testing sets. Short scanpaths with a sequence length less than $4$ are discarded, following~\cite{martin2024tspm,sun2019visual}, and all fixation scanpaths are padded to a maximum sequence length of $8$, similar to the work in~\cite{martin2024tspm}. The padded fixations in the scanpath are discarded when computing the loss. 

The MIT Low Resolution dataset~\cite{judd2011fixations} was used to test the model performance across different datasets, following a similar evaluation procedure as~\cite{sun2019visual,martin2024tspm}. The MIT Low Resolution dataset~\cite{judd2011fixations} contains $168$ natural images at $8$ different resolution. We only consider the highest resolution group following~\cite{sun2019visual}. Each image is viewed by $8$ subjects.

% The input image is resized to $256\times384$ before feeding it to
% the model. The downsampling $S$ is set to $4$ for GazeLNN and $8$ for GazeLNN-S. The fixation scanpath length is fixed to $8$.
% The model was trained for $100$ epochs using Adam op-
% timize with learning rate of $0.0001$. The learning rate was
% kept constant throughout training.
The input image is resized to $256\times 384$, following~\cite{martin2024tspm}, before feeding it to the model. The downsampling $S$ is set to $8$, resulting in a reduced spatial representation that remains sufficiently high-dimensional for processing by the fully connected layers in the CfC module. 

The fixation scanpath length is fixed to $8$, following~\cite{martin2024tspm}. The fixation scanpath is converted into a single fixation heatmap before integration with the active camera control policy, aggregating the sequence of fixation points into a continuous spatial representation of visual attention. The elapsed time $\Delta t$ in the \ac{cfc} module is obtained from the ground-truth fixation durations, and for the first fixation it is set to zero. During deployment with the active camera policy, $\Delta t$ is fixed to $1$ due to the absence of ground-truth duration; this value corresponds to the default setting in the~\ac{cfc} PyTorch implementation.

% The fixation scanpath length is fixed to $8$, following~\cite{martin2024tspm}. The scanpath is converted into a single fixation heatmap representing the spatial distribution of visual attention. The elapsed time $\Delta t$ in the \ac{cfc} module is derived from ground-truth fixation durations (set to $0$ for the first fixation). During deployment, $\Delta t$ is fixed to $1$ due to the absence of ground-truth timing, following the default setting in the~\ac{cfc} PyTorch implementation.

The model was trained for $100$ epochs using the Adam optimizer with a learning rate of $0.0001$. The learning rate was kept constant throughout training. The proposed Dynamic Time Wrapping with KL-Divergence loss (KL-DTW) in~\cite{martin2024tspm} was used to optimize the whole scanpath while accounting for both the spatial and temporal nature of the fixation scanpath. Early stopping was applied based on validation with patience for $20$ epochs to prevent overfitting. All experiments were conducted on a laptop equipped with an NVIDIA RTX 3500 Ada GPU. Inference time is reported as the average per-image latency across the test set (excluding the first image to avoid warm-up overhead).

\section{REINFORCEMENT LEARNING FOR FIXATION-GUIDED ACTIVE PERCEPTION}
\label{sec:active_perception_rl}
To demonstrate the utility of the predicted scanpaths for autonomous robot navigation, we integrate GazeLNN into an active camera-robot control policy trained via \ac{rl}. Building upon the framework proposed in~\cite{malczyk2026reinforcement}, the policy jointly optimizes goal-directed navigation, collision avoidance, and human-fixation-guided behavior. It achieves this by directing an actuated onboard camera toward regions predicted to attract human fixations. To train this policy, we utilize the Aerial Gym simulator~\cite{kulkarni2025aerial}, which provides the depth and segmentation streams required to simulate visual attention. The policy receives observations comprising the robot state (position, orientation, linear and angular velocities), the current depth image and its associated fixation heatmap, a local ego-centric 3D occupancy grid encoding nearby free and occupied space, and the current camera orientation (pitch and yaw). The policy outputs an action vector commanding the robot's linear velocities and yaw rate, alongside the desired pitch and yaw angles of the actuated camera, as shown in~\Cref{fig:network_architecture}. The action vector reads:
\begin{align}
    \mathbf{a}_t = [\underbrace{{\mathbf{v}_t^{r}},~{\omega}_{t,z}^{r}}_{\mathbf{a}^\textrm{nav}_t}, ~\underbrace{{\mathbf{c}_t^{r}}}_{\mathbf{a}^{\textrm{cam}}_t}],
\label{eq:action}
\vspace{-5ex}
\end{align}
where $\mathbf{v}_t^{r}\in \mathbb{R}^3$ and ${\omega}_{t,z}^{r}$ are the commanded linear velocities and yaw rate expressed in vehicle frame, and $\mathbf{c}^r_{t} = \{{\chi}_{t}^{r}, {\psi}_{t}^{r}\}$ denotes the commanded pitch and yaw angles for the camera.

\subsection{Heatmap Generation During Training}

Since GazeLNN is deployed for live predictions during inference, we must generate proxy fixation heatmaps to facilitate \ac{rl} training within the simulator. The process begins by sampling the face-mesh indices of simulated obstacle meshes and reducing the segmented mesh-faces to a set of representative fixation heatmap points. Each fixation point is then randomly perturbed by up to $5$ pixels to introduce spatial noise and improve policy robustness. Finally, the sparse point set is convolved with a Gaussian kernel to produce smooth heatmaps consistent with the output format of GazeLNN.

\subsection{Reward Function}
Expanding upon the approach in~\cite{malczyk2026reinforcement}, the reward function combines navigation progress, motion smoothness, collision avoidance, and a novel \emph{fixation-attraction} term:
\begin{equation}
    R(s_t, a_t) = r_t + l_t + p_t + h_t.
\end{equation}
Specifically, $r_t$ rewards navigation progress toward the target waypoint, defined as the reduction in Euclidean distance $d_t$ to the goal: 
\begin{align}
    r_t = w_r (d_{t-1} - d_t),
\end{align}
with $w_r$ defined as a scaling weight. The smoothness penalty $l_t$ discourages erratic, jerky movements of both the robot base and the camera by penalizing large variations in consecutive actions $\mathbf{a}_t = \big[{\mathbf{a}^{\mathrm{nav}}_t}, \, {\mathbf{a}^{\mathrm{cam}}_t}\big]$:
\begin{align}
    l_t = -w_l \|\mathbf{a}_t - \mathbf{a}_{t-1}\|^2,
\end{align}
where $w_l>0$ is a scaling motion smoothness weight. The collision avoidance term $p_t$ penalizes proximity to obstacles using the local occupancy grid, applying a negative reward $-w_p$ if the distance to the nearest obstacle falls below a predefined safety threshold calculated based on the robot size.
% where $r_t$ rewards navigation progress toward the goal, $l_t$ penalizes jerky motion of both the robot and the camera, and $p_t$ penalizes proximity to obstacles using the local occupancy grid. These three terms follow directly from~\cite{malczyk2026reinforcement}. The key contribution of this work is the \emph{fixation-attraction} term $h_t$, which
% replaces the voxel-based information gain of~\cite{malczyk2026reinforcement} and incentivizes the camera to fixate on regions predicted to be visually salient by GazeLNN. Given the fixation heatmap $\mathbf{H}_t$ at timestep $t$, we define:
To introduce a human-fixation-guided behavior, we introduce the \emph{fixation-attraction} term $h_t$, which replaces the voxel-based information gain used in prior work and incentivizes the camera to fixate on regions predicted to be visually salient by GazeLNN. Given the fixation heatmap $\mathbf{H}_t$ at timestep $t$, we define:
\begin{equation}
    h_t = w_h \frac{\displaystyle\sum_{u,v} \mathbf{H}_t(u,v)\cdot\exp\!\left(-\alpha\, d(u,v)^2\right)}
               {\displaystyle\sum_{u,v} \mathbf{H}_t(u,v)>0},
    \label{eq:attraction}
\end{equation}
where $d(u,v)$ is the Euclidean pixel distance from the image center, $\alpha>0$ is a spatial decay parameter, $w_h>0$ is a scaling attention weight, and the denominator normalizes by the number of active heatmap pixels to prevent the reward from scaling with fixation density. Jointly, $h_t$ steers the policy to continuously reorient the camera so that predicted fixation-attracting regions are kept near the center of the camera's field of view.

\begin{figure}[t]
    \centering
    \includegraphics[width=0.99\columnwidth]{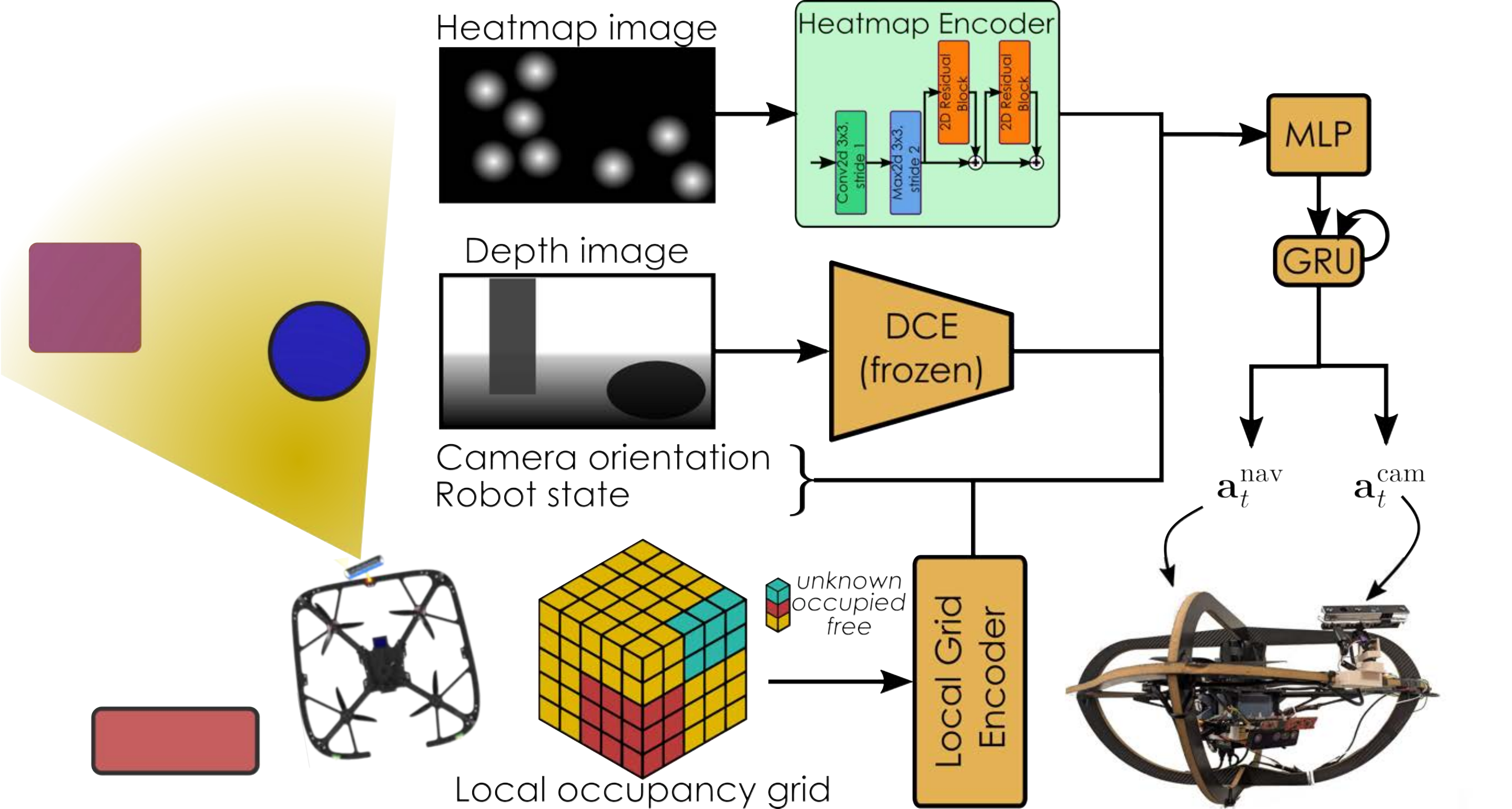}
    \vspace{-4ex}
    \caption{Proposed policy network architecture. A fixation heatmap encoder and a frozen depth encoder process the visual inputs, whose latent features are concatenated with local occupancy grid features, robot state, and camera orientation, and passed through an MLP and GRU to produce navigation and camera actions.}
    \label{fig:network_architecture}
    \vspace{-4ex}
\end{figure}

\subsection{Implementation}
The policy is trained using the \ac{appo} algorithm~\cite{petrenko2020sample} within Aerial Gym, utilizing established environments and domain randomization schemes. Prior to encoding, both the depth and fixation heatmap images are downsampled to $96\times54$ pixels. This reduces computational overhead while maintaining adequate spatial resolution for fixation-guided camera control and collision avoidance. For feature extraction, we introduce a dedicated heatmap encoder based on a 2D ResNet architecture~\cite{espeholt2018impala,malczyk2025semantically}. Simultaneously, the depth image representation is compressed using a pre-trained and frozen \ac{dce}~\cite{kulkarni2023task}.

\begin{table*}[t]
\centering
% \caption{Comparison between our proposed model GazeLNN and GazeLNN-S against different state-of-the-art recurrent-based models and Human baseline as reported in~\cite{martin2024tspm}. The best performance is highlighted in bold, while the second best is underlined. }
\caption{Comparison between our proposed model GazeLNN against different state-of-the-art recurrent-based models and Human baseline as reported in~\cite{martin2024tspm}. The best performance is highlighted in bold, while the second best is underlined. }
\vspace{-2ex}
\label{tab:comparison against state-of-the art methods}
\resizebox{\textwidth}{!}{%
\begin{tabular}{lcccccc}
\toprule
Model &
\multicolumn{2}{c}{String alignment} &
\multicolumn{2}{c}{Curve similarities} &
\multicolumn{2}{c}{Time-series analysis} \\
\cmidrule(lr){2-3} \cmidrule(lr){4-5} \cmidrule(lr){6-7}
 & Lev. Dist. $\downarrow$ 
 & ScanMatch $\uparrow$
 & Hausdorff Dist. $\downarrow$
 & Frechet Dist. $\downarrow$
 & fast DTW $\downarrow$
 & T. D. Emb. $\downarrow$ \\
\midrule
Human BL  & 10.77 (1.61) & 0.38 (0.06) & 95.97 (18.40) & 140.02 (26.16) & 550.84 (133.71) & 42.40 (8.45) \\
\midrule
Itti Model~\cite{itti2002model}   & 14.04 (0.80) & 0.23 (0.05) & 160.09 (29.31) & 207.97 (27.21) & 1041.16 (153.97) & 63.88 (9.54) \\
LeMeur Model~\cite{le2016introducing} & 12.58 (0.78) & \underline{0.35} (0.04) & 104.84 (12.79) & 163.59 (20.52) & 669.67 (108.49) & 39.75 (6.53) \\
IOR-ROI~\cite{sun2019visual}       & 13.26 (0.71) & 0.30 (0.05) & 115.50 (20.22) & 166.07 (21.69) & 777.75 (119.46) & 46.98 (7.18) \\
Chen Model~\cite{chen2018scanpath}   & 13.04 (1.14) & 0.31 (0.07) & 109.18 (27.38) & 149.32 (33.03) & 682.80 (183.11) & 46.90 (12.55) \\
tSPM-Net~\cite{martin2024tspm}&  \underline{11.47} (1.13) & 0.34 (0.06) & \underline{103.44} (27.13) & \underline{144.77} (32.77) & \underline{610.02} (155.96) & \underline{43.74} (10.25) \\
% \textbf{GazeLNN}& \textbf{11.38} (2.47)  & \textbf{0.46} (0.11) & \underline{107.27}(50.97) & \textbf{136.03} (53.70)  & \textbf{536.66} (199.04)& \textbf{39.75} (18.81)  \\
% \textbf{GazeLNN-S}& 11.61 (2.58)  & \underline{0.43} (0.13) & 115.67 (53.67) & \underline{144.60} (54.68)  & 621.39 (244.74)& \underline{42.81} (16.39)  \\
\textbf{GazeLNN (ours)}& \textbf{11.22} (2.57) & \textbf{0.47} (0.11) & \textbf{98.17} (47.17) & \textbf{133.31} (53.27) & \textbf{537.72} (37.85) & \textbf{27.20 }(19.29) \\
\bottomrule
\end{tabular}
}
\vspace{-1ex}
\end{table*}

As illustrated in~\Cref{fig:network_architecture}, the resulting 2D latent features are concatenated with embeddings from the 3D occupancy grid encoder, the robot state, and the camera orientation. These combined features are processed by a \ac{mlp} and subsequently passed through a \ac{gru} block. The recurrent unit enables the temporal integration of observations, which is crucial for handling partial observability when obstacles or salient regions temporarily leave the sensor frustum. Finally, the policy outputs a six-dimensional action vector that jointly commands the robot's navigation ${\mathbf{a}^{\mathrm{nav}}_t}$ and the camera's orientation ${\mathbf{a}^{\mathrm{cam}}_t}$, enabling safe, goal-directed navigation coupled with human-like visual attention.

% As illustrated in~\Cref{fig:network_architecture}, the resulting 2D latent features are concatenated with embeddings from the 3D occupancy grid encoder~\cite{malczyk2026reinforcement}, the robot state, and the camera orientation. These combined features are processed by an \ac{mlp} and subsequently passed through a \ac{gru} block. The recurrent unit enables the temporal integration of observations, which is crucial for handling partial observability when obstacles or salient regions temporarily leave the sensor frustum. Finally, the policy outputs a six-dimensional action vector that jointly commands the robot's navigation ${\mathbf{a}^{\mathrm{nav}}_t}$ and the camera's orientation ${\mathbf{a}^{\mathrm{cam}}_t}$, enabling safe, goal-directed navigation coupled with human-like visual attention based on the GazeLNN model.

\section{GazeLNN EVALUATION}
\label{sec: results}
This section provides an overview of the metrics used in fixation scanpath assessment, followed by a comparison with the state-of-the-art methods. An ablation study on the choice of the backbone and the effect of different types of recurrent networks is also provided.
\subsection{Metrics}
\label{sec:metrics}
Following~\cite{martin2024tspm}, we evaluate scanpaths using the metrics proposed in~\cite{fahimi2021metrics}, including string alignment metrics (Levenshtein Distance, ScanMatch), curve similarity metrics (Hausdorff Distance, Fréchet Distance), and time-series metrics (\ac{dtw}, \ac{tde}). For string-based metrics, the image is discretized using a $12\times8$ grid, converting fixations into symbolic sequences based on grid cells. Levenshtein Distance measures sequence differences without weighted costs, while ScanMatch applies a weighted substitution cost based on Euclidean distance with a similarity threshold of $3.5$~\cite{fahimi2021metrics,martin2024tspm}. Hausdorff Distance measures the maximum spatial mismatch between scanpaths, whereas Fréchet Distance preserves fixation ordering when comparing trajectories. \ac{dtw} aligns scanpaths of varying lengths under boundary, continuity, and monotonicity constraints, while~\ac{tde} captures temporal structure by comparing subsequences of fixations.
\subsection {Comparison against state-of-the art methods}
\label {sec:Comparison against state-of-the art methods}

\Cref{tab:comparison against state-of-the art methods} compares GazeLNN against state-of-the-art recurrent-based methods. The results for all baseline models are directly reported from~\cite{martin2024tspm}, while the performance of GazeLNN is obtained using the same dataset and evaluation protocol. Recurrence-based metrics are excluded because they are sensitive to the choice of recurrence threshold, which is not specified in~\cite{martin2024tspm}. Overall, GazeLNN achieves superior performance across all of the reported metrics. In terms of computational costs, GazeLNN reduces the number of parameters and floating point operations significantly compared to tSPM-Net (second best model) by $91.80\%$ trainable parameters and $99.40\%,$ GFLOPS respectively. GazeLNN provides higher inference speed with a speedup of $6.42$ times and compared to tSPM-Net.

% Figure~\ref{fig:scanpath visualization} presents a qualitative comparison of the predicted scanpaths by both variants of GazeLNN and tSPM-Net~\cite{martin2024tspm} (the second-best model) against ground truth scanpath. The predicted scanpath by GazeLNN model closely resembles the human ground truth scanpath compared to tSPM-Net~\cite{martin2024tspm}, while GazeLNN-S tends to provide less exploratory scanpath with reduced spatial coverage.

\begin{figure*}[t]
    \centering
    \setlength{\tabcolsep}{4pt}
    \begin{tabular}{ccccc}
        Ground Truth &
        \raisebox{-0.5\height}{\includegraphics[width=0.2\textwidth]{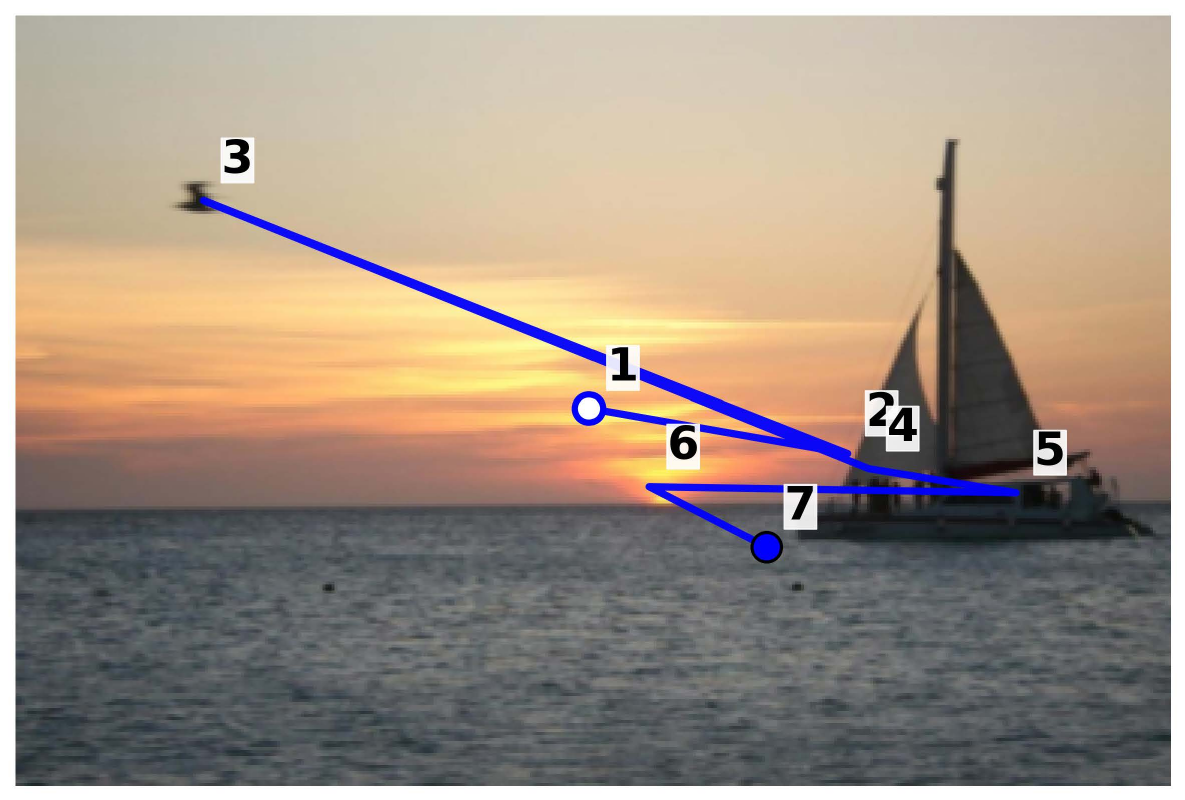}} &
        \raisebox{-0.5\height}{\includegraphics[width=0.2\textwidth]{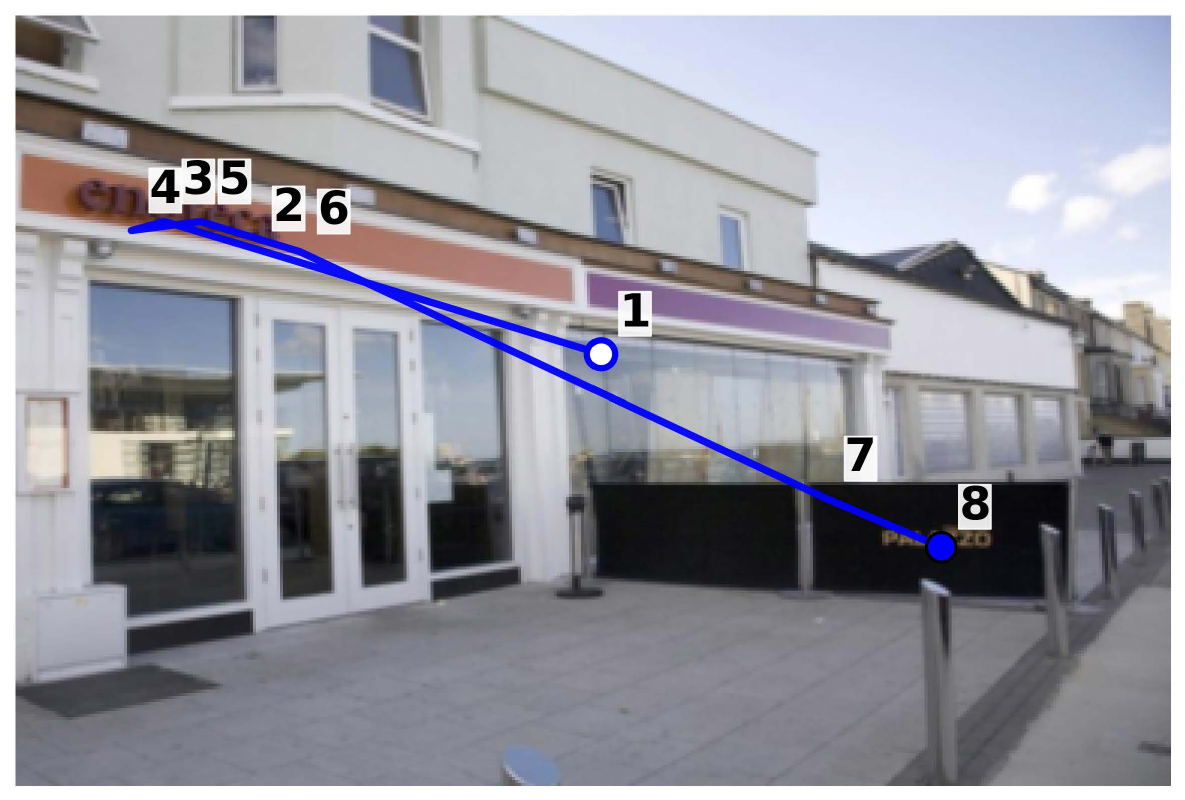}} &
        \raisebox{-0.5\height}{\includegraphics[width=0.2\textwidth]{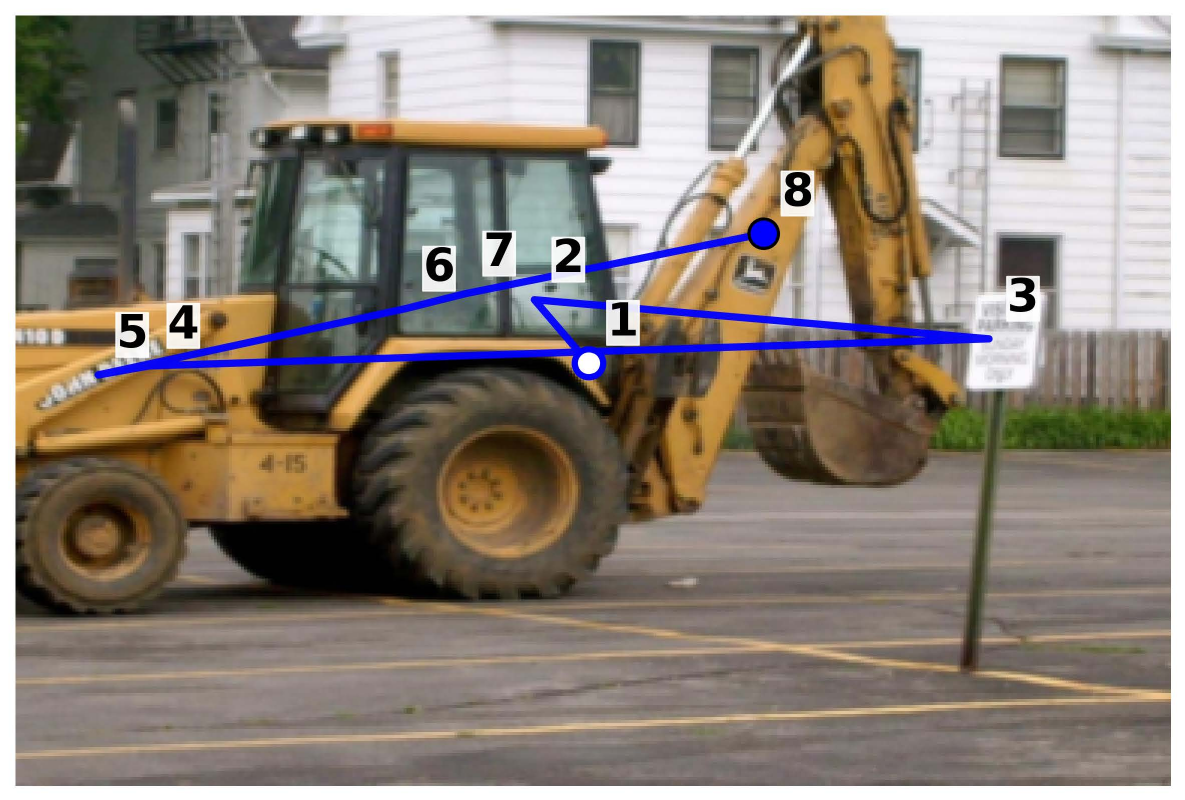}} 
         \raisebox{-0.5\height}{\includegraphics[width=0.2\textwidth]{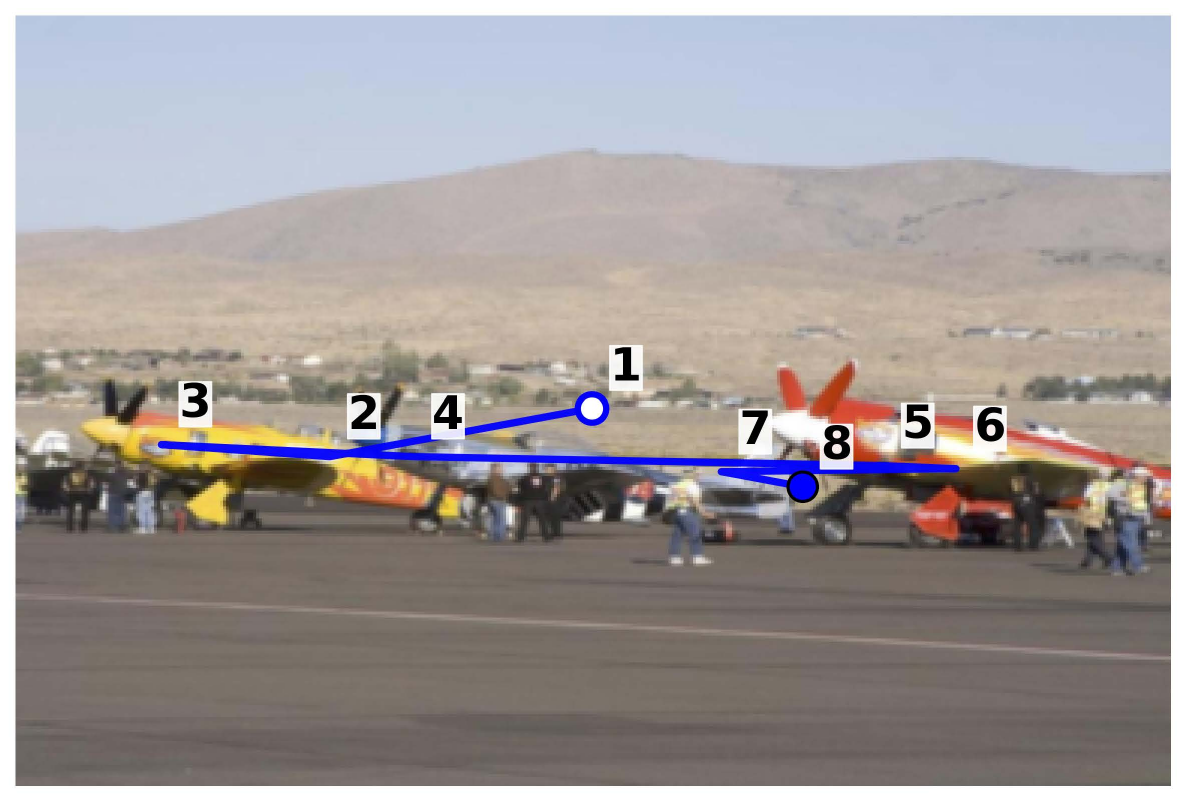}} \\

        tSPM-Net &
        \raisebox{-0.5\height}{\includegraphics[width=0.2\textwidth]{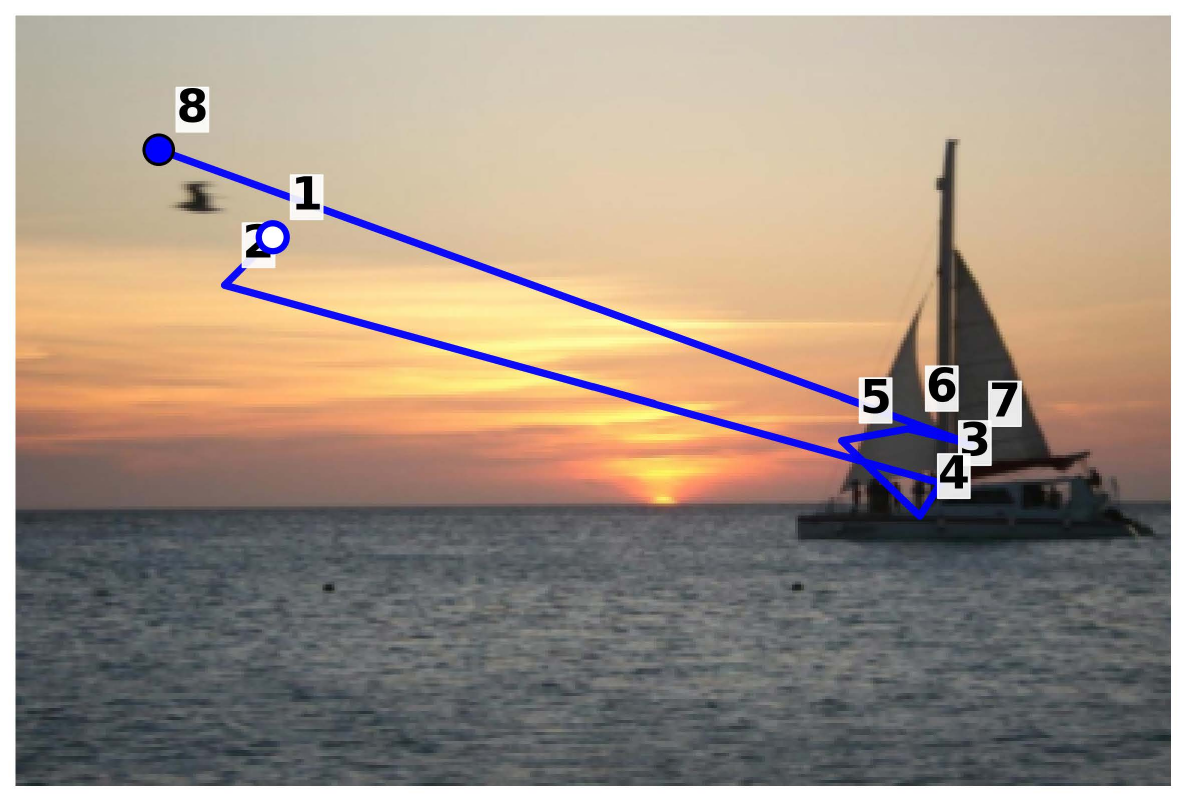}} &
        \raisebox{-0.5\height}{\includegraphics[width=0.2\textwidth]{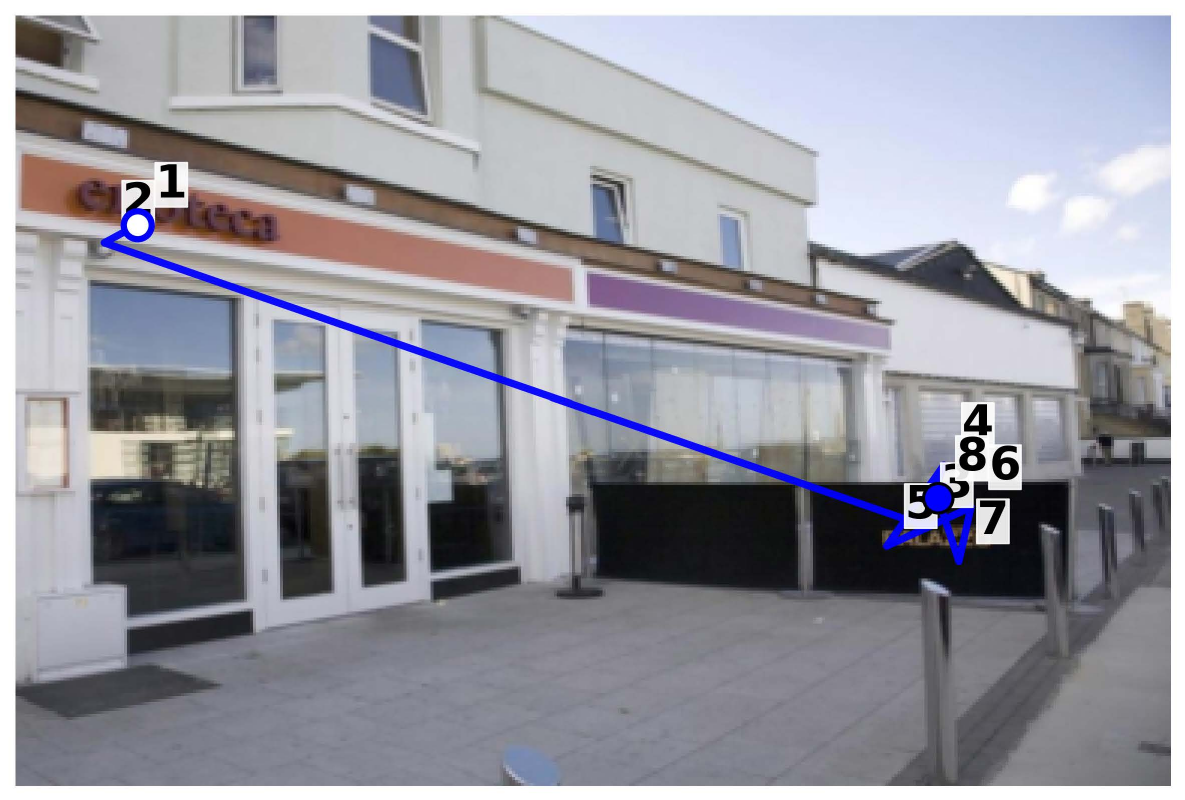}} &
        \raisebox{-0.5\height}{\includegraphics[width=0.2\textwidth]{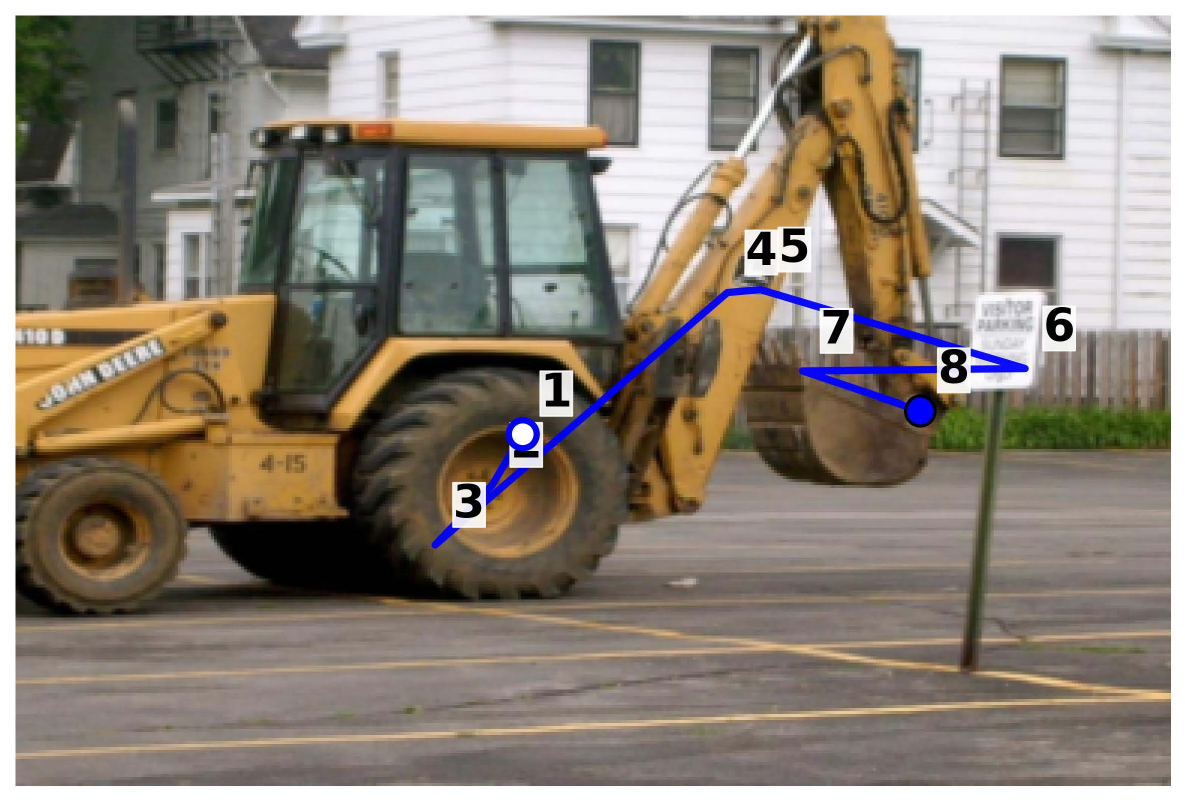}} 
        \raisebox{-0.5\height}{\includegraphics[width=0.2\textwidth]{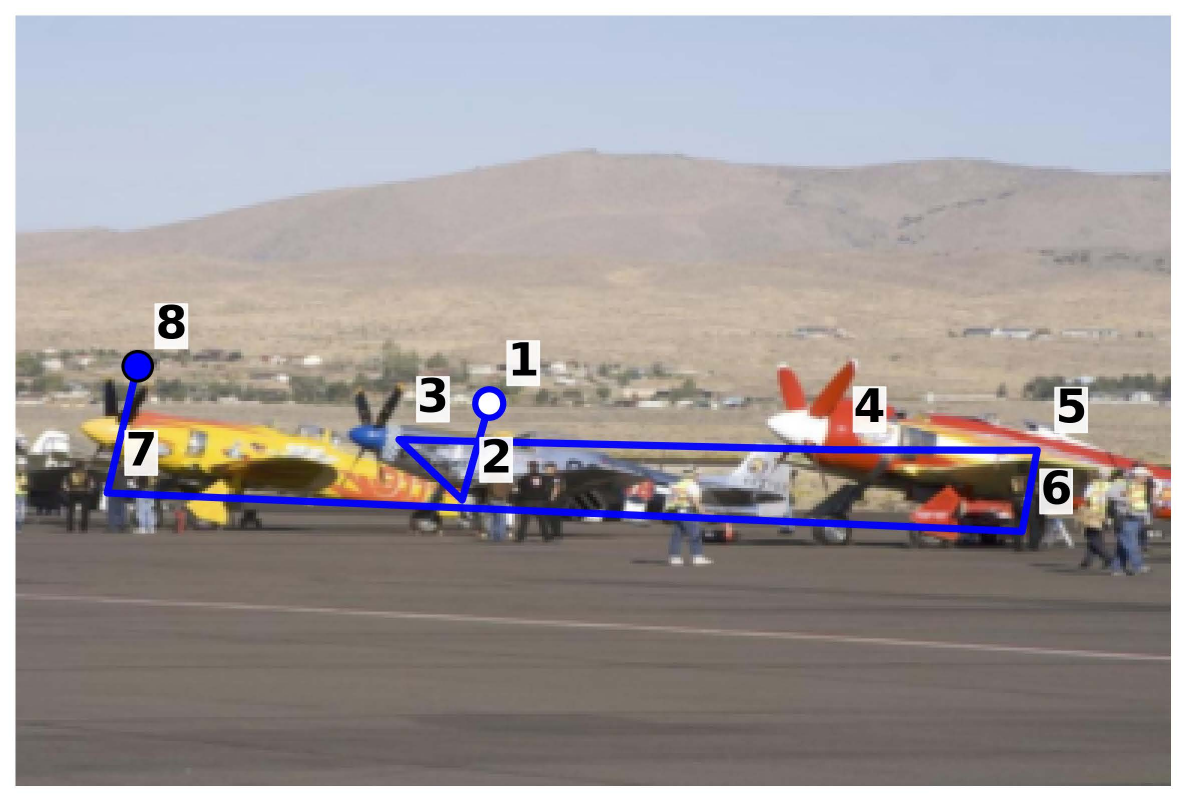}} \\

        GazeLNN &
        \raisebox{-0.5\height}{\includegraphics[width=0.2\textwidth]{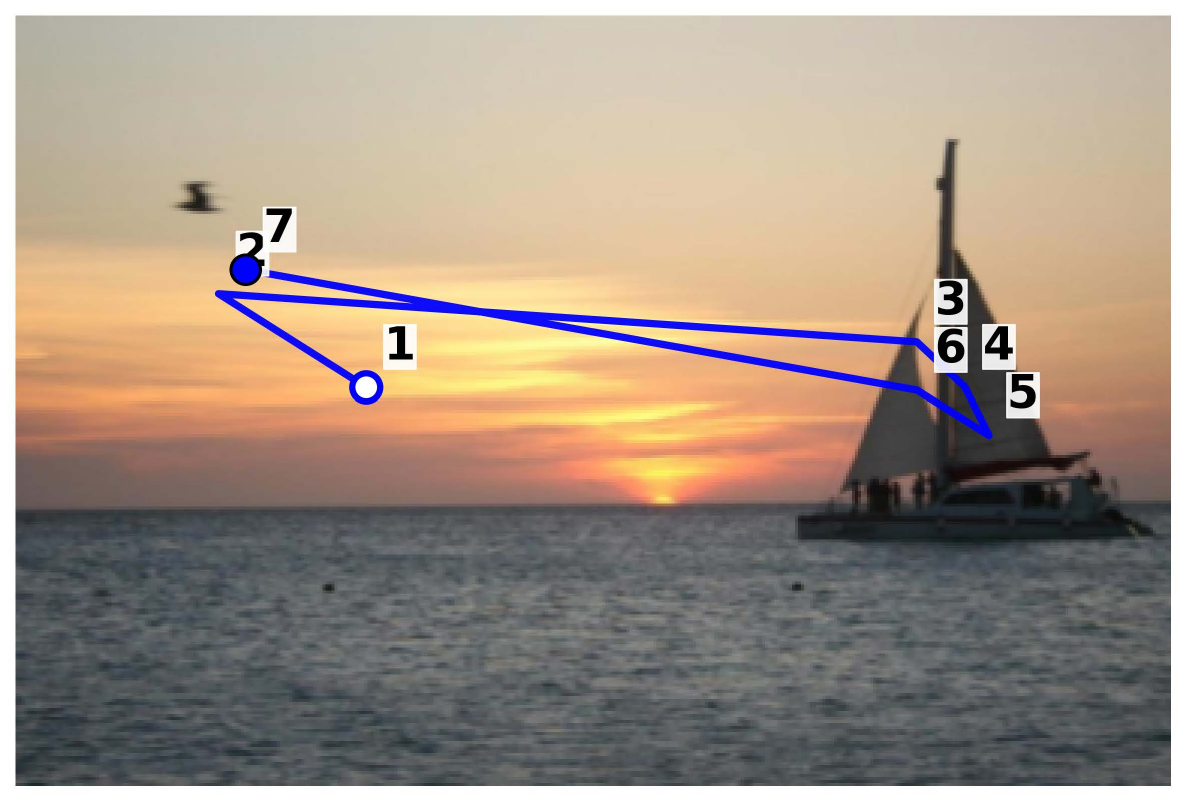}} &
        \raisebox{-0.5\height}{\includegraphics[width=0.2\textwidth]{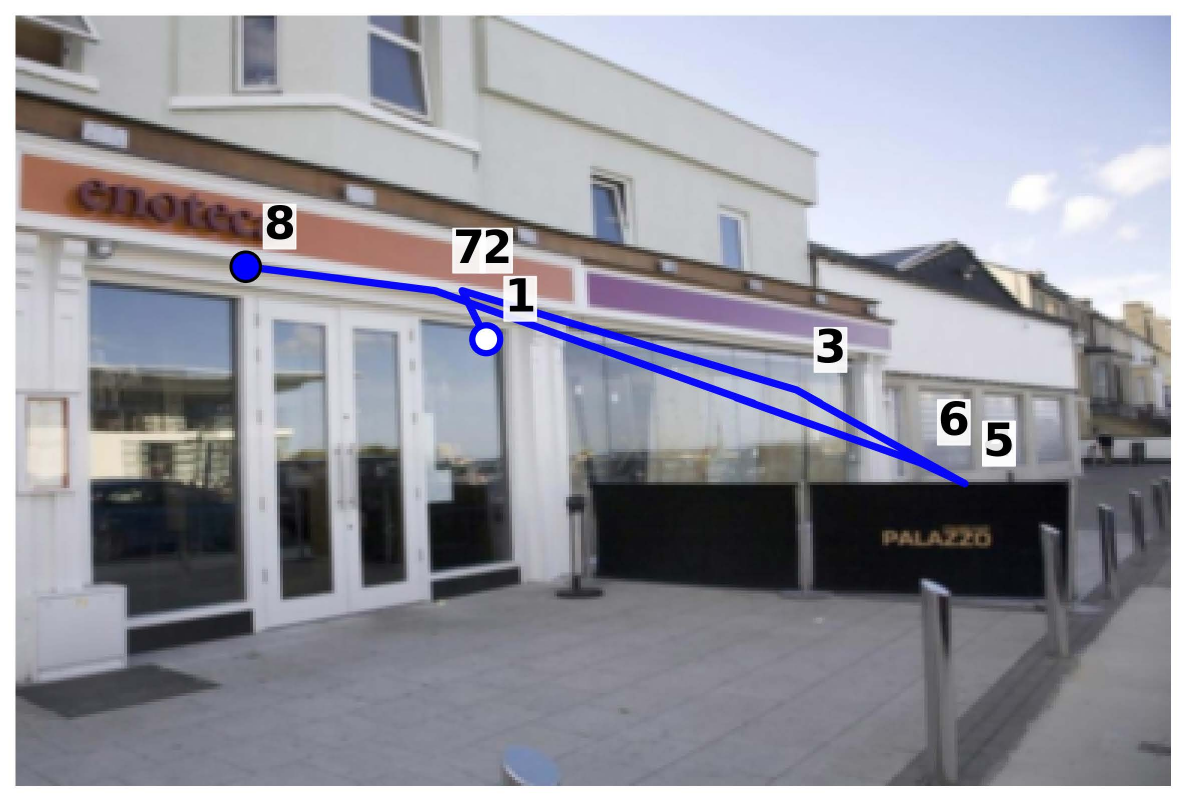}} &
        \raisebox{-0.5\height}{\includegraphics[width=0.2\textwidth]{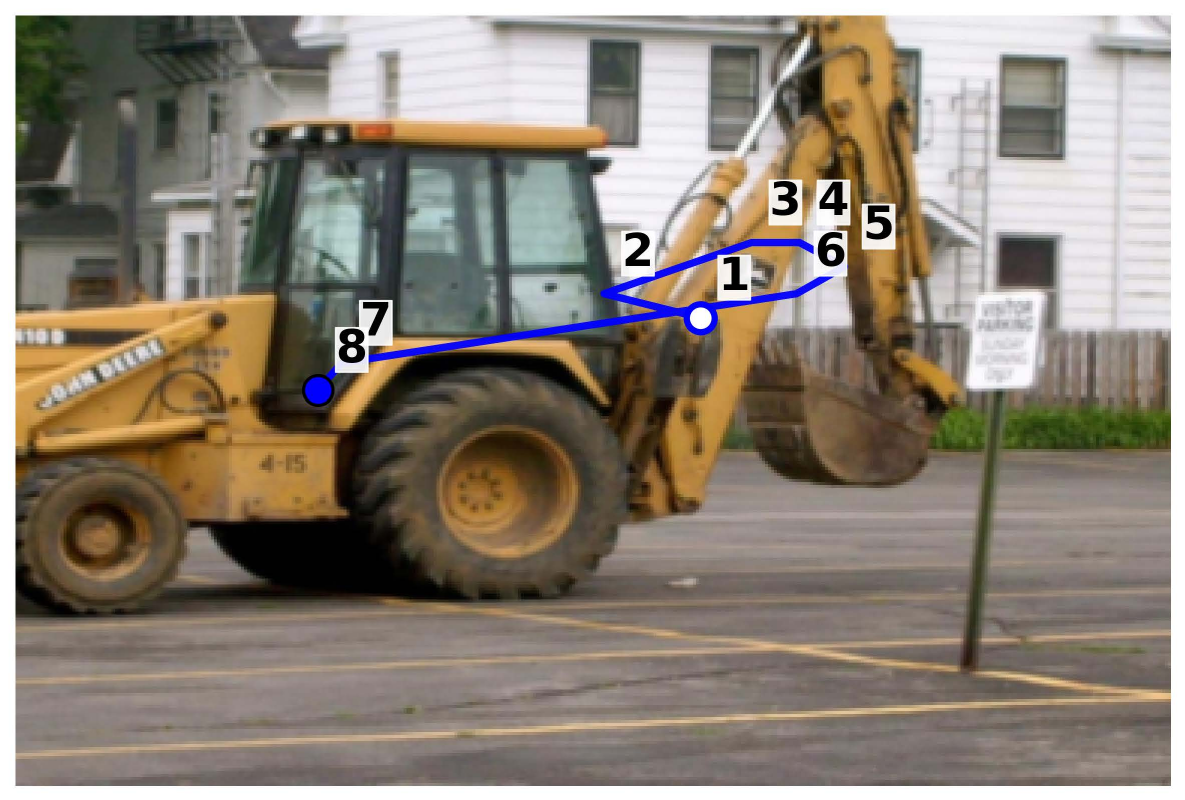}} 
        \raisebox{-0.5\height}{\includegraphics[width=0.2\textwidth]{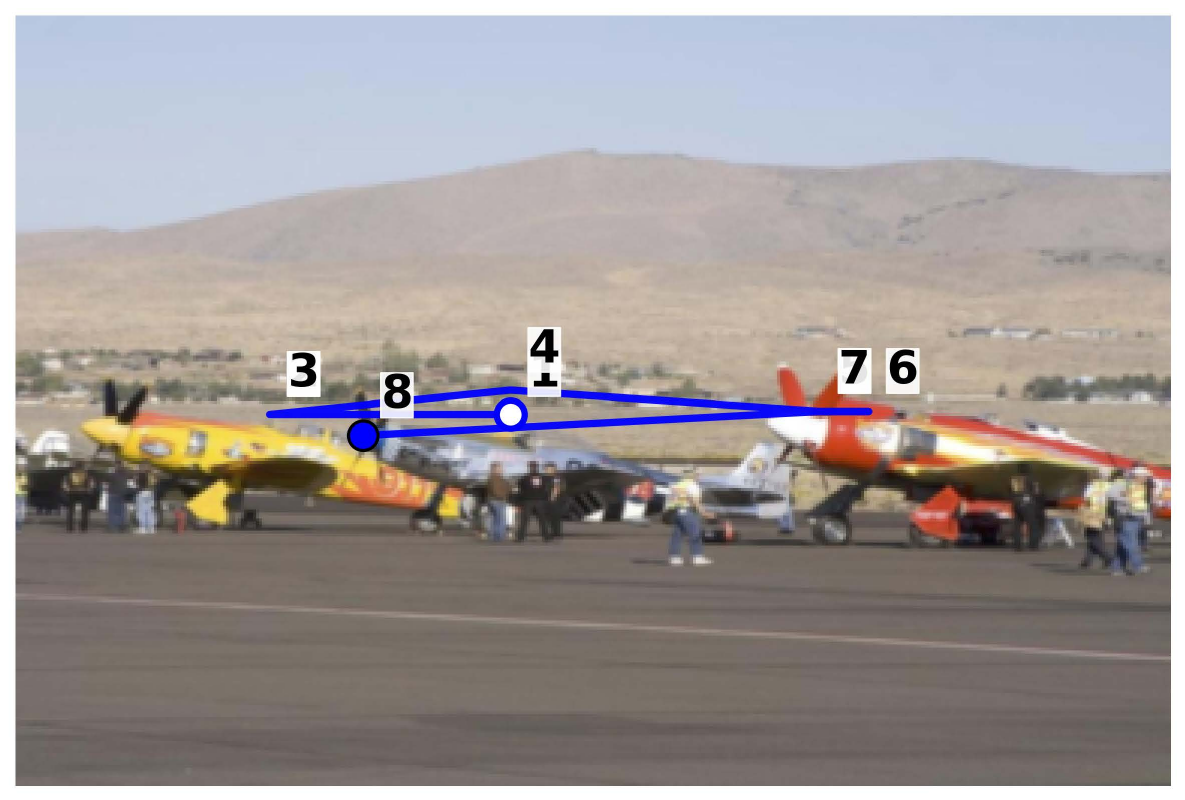}} \\

    \end{tabular}
    \vspace{-1ex}
    \caption{Scanpath visualizations for GazeLNN compared to tSPM-Net for a small set of images (qualitative analysis).}
    \label{fig:scanpath_visualization}
    % \vspace{-2ex}
\end{figure*}

Figure~\ref{fig:scanpath_visualization} presents a qualitative comparison of the predicted scanpaths by GazeLNN and tSPM-Net~\cite{martin2024tspm} (the second-best model) against the ground truth scanpath. The predicted scanpath by GazeLNN model more closely resembles the human ground truth scanpath as compared to tSPM-Net~\cite{martin2024tspm}.

% \begin{figure}[t]
%     \centering
%     \setlength{\tabcolsep}{4pt} % Adjust column spacing
%     \begin{tabular}{ccc}
%         Ground  &  GazeLNN & tSPM-Net \\ 
%          \raisebox{-0.5\height}{\includegraphics[width=0.30\columnwidth]{Figures/results/i169636965_lowRes512_subjectid_bwk_GT.png}} &
%         \raisebox{-0.5\height}{\includegraphics[width=0.30\columnwidth]{Figures/results/i169636965_lowRes512_subjectid_elp_Prediction_mobilenet_updated.png}} & 
%         \raisebox{-0.5\height}{\includegraphics[width=0.30\columnwidth]{Figures/results/i169636965_lowRes512_3_tsmp.png}} \\
%          \raisebox{-0.5\height}{\includegraphics[width=0.30\columnwidth]{Figures/results/i1733711537_lowRes512_subjectid_iii_GT.png}} & 
%         \raisebox{-0.5\height}{\includegraphics[width=0.30\columnwidth]{Figures/results/i1733711537_lowRes512_subjectid_khe_Prediction_mobilenet_updated.png}} & 
%         \raisebox{-0.5\height}{\includegraphics[width=0.30\columnwidth]{Figures/results/i1733711537_lowRes512_tssmp_6.png}}  \\
%         \raisebox{-0.5\height}{\includegraphics[width=0.30\columnwidth]{Figures/results/i2145105890_lowRes512_subjectid_jul_GT.png}} &
%         \raisebox{-0.5\height}{\includegraphics[width=0.30\columnwidth]{Figures/results/i2145105890_lowRes512_subjectid_mpw_Prediction_mobilenet_updated.png}} & 
%         \raisebox{-0.5\height}{\includegraphics[width=0.30\columnwidth]{Figures/results/i2145105890_lowRes512_2_tsmp.png}} \\
%     \end{tabular}
    
%       \caption{Scanpath visualizations for GazeLNN compared to tSPM-Net. } 
%        \label{fig:scanpath visualization}
% \end{figure}

\subsection{Backbone Ablations}
\label{sec: backbone ablation}
We evaluate the impact of the feature extraction backbone by comparing different alternative architectures. Specifically, we begin with the backbone configuration used in tSPM-Net~\cite{martin2024tspm}, which combines VGG19 with a DeepLabV3~\cite{yurtkulu2019semantic} segmentation module. We first replace VGG19~\cite{simonyan2014very} with ResNet50 as it has shown better perfomance with reduced computational costs~\cite{he2016deep}, and with MobileNetV3~\cite{howard2019searching} to obtain a more lightweight variant. In addition, we examine the effect of removing the DeepLabV3 module to further reduce the complexity of the model.

% We further investigate different variants of DINOv3, given its strong pretrained feature representations and demonstrated performance across diverse vision tasks. Particularly, we evaluate the original DINOv3 backbone and its distilled small variant (ViT-S), which provides a significantly lighter architecture while maintaining competitive representational capacity. Finally, we assess the effect of projecting and upsampling the DINOv3 (ViT-S) features before feeding them into the Conv-CfC cell, denoted as DINOv3 (ViT-S)*.

% We further investigate different variants of DINOv3, given its strong pretrained feature representations and demonstrated performance across diverse vision tasks. Particularly, we evaluate the original DINOv3 backbone and its distilled small variant (ViT-S), which provides a significantly lighter architecture while maintaining competitive representational capacity. Finally, we assess the effect of projecting and upsampling the DINOv3 (ViT-S) features before feeding them into the CfC cell, denoted as DINOv3 (ViT-S)*.
We further investigate variants of DINOv3~\cite{simeoni2025dinov3} due to its strong pretrained representations and performance across vision tasks. Specifically, we evaluate the original DINOv3 backbone and its distilled small variant (ViT-S), which provides a lighter architecture while maintaining competitive capacity. Finally, we examine projecting and upsampling the DINOv3 (ViT-S) features before feeding them into the CfC cell, denoted as DINOv3 (ViT-S)*.

For all the above-mentioned backbones, the pretrained models were finetuned during training on the OSIE dataset, except for (a) all DINOv3 backbones which were frozen, and (b) the first half of VGG19 which was frozen as described in~\cite{martin2024tspm}. For the model variants using DeepLabV3, the extracted segmentation masks are concatenated with the features extracted by the other backbone and projected to a single feature map following the same procedure in tSPM-Net~\cite{martin2024tspm}, where the concatenated features are passed to two convolution layers with kernel size of $1\times1$ and output channels of $64$ and $1$. Each convolution layer is followed by a tanh activation.

Table~\ref{tab:backbone comparsion} summarizes the comparison between the above-mentioned backbone variants in terms of performance (ScanMatch score) and computational costs. The ScanMatch metric is chosen as it is a widely used metric in the literature~\cite{martin2024tspm,chen2018scanpath,sun2019visual,le2016introducing}. The rest of the model is kept unchanged across all comparisons. 

% DINOv3(VIT-S)* achieves the highest performance whereas VGG19+DeepLabV3 has the lowest performance. Resnet50 and MobilenetV3 variants with DeepLabV3 enhances the model performance, while significantly reducing the model parameters and floating point operations, compared to the VGG19 variant with DeeplabV3. DeepLabV3 module doesn't have a significant impact on the model performance as shown in the case of using Resnet50 and Mobilenet solely, while removing it results in even more lighter weight model. The most efficient variant, in terms of computational cost and speed, is MobilenetV3 as it is designed for mobile and embedded devices~\cite{howard2019searching}.

Overall, all backbones achieve similar performance on the ScanMatch score, while the major difference is in computational cost. The backbone variants without DeeplabV3 provide less inference time and reduced computational costs. Among all evaluated backbones, MobileNetV3 is the most efficient in terms of computational cost and inference speed, as it is specifically designed for mobile and embedded devices~\cite{howard2019searching}. Therefore, MobileNetV3 is selected as the backbone for our model.

\begin{table}[t]
\vspace{-2ex}
\centering
\caption{Backbone comparison: performance vs. computational cost.}
\vspace{-2ex}
\label{tab:backbone comparsion}
\resizebox{\columnwidth}{!}{%
\large
\begin{tabular}{lcccc}
\toprule
\textbf{Backbone} &
 \textbf{ScanMatch $\uparrow$}
& \textbf{Time (ms)$\downarrow$}
& \textbf{\#Params (M)$\downarrow$}
& \textbf{GFLOPs}$\downarrow$ \\
\midrule
VGG19+DeepLabV3    & 0.47 (0.10)  & 17.43 & 195.41 & 99.81 \\
Resnet50+DeepLabV3   & 0.46 (0.10) & 14.32 & 77.4 & 69.8 \\
MobilenetV3+DeepLabV3 & \textbf{0.48} (0.11)  & 13.80 &  57.25 & 62.07 \\
Resnet50   & 0.46 (0.11)  & 7.39  & 35.39 & 8.33 \\
MobilenetV3  & 0.47 (0.11) & \textbf{6.84} & \textbf{15.24} & \textbf{0.61} \\
DINOv3  & \textbf{0.48} (0.11)) & 15.65  &  90.15 & 33.38 \\
DINOv3(VIT-S) &  0.47 (0.11) & 10.02 & 26.04 & 8.45 \\
DINOv3(VIT-S)*  & \textbf{0.48} (0.11)  & 8.76 &  31.35 & 8.49\\
\bottomrule
\end{tabular}
}
\vspace{-5ex}
\end{table}

\subsection{Recurrent network ablation}
\label{sec:rnn ablation}
% We study the effect of different recurrent network architectures on model performance and computational costs. Specifically, we compare the Bayesian ConvLSTM used in tSPM-Net~\cite{martin2024tspm}, against the normal ConvLSTM (with the same configuration as the Bayesian ConLSTM) and our proposed Conv-CfC variant of the liquid neural network. In all experiments, the VGG19+DeeplabV3 backbone is used similar to tSPM-Net~\cite{martin2024tspm}, while keeping all the training setup unchanged. Table~\ref{tab:rnn comparsion} summarizes the comparison between different RNN architectures, where the ScanMatch scores for Bayesian ConvLSTM and ConvLSTM are reported as in~\cite{martin2024tspm}. Conv-CfC is the most efficient variant in terms of computational costs, while having better performance compared to the ConvLSTM based variants. More specifically the Conv-CFC module has 4.55K paramaters compared to 0.21M, and 53.48K for Bayesian ConvLSTM and ConvLSTM, respectively. Bayesian ConvLSTM requires mean and variance resulting in doubling the number of parameters compared to normal ConvLSTM. In terms of computational cost, Conv-CfC module requires only 0.05 GFLOPS whereas Bayesian ConvLSTM and ConvLSTM require 2.62 GFLOPS.

 We study the effect of different recurrent network architectures commonly used in the scanpath prediction literature~\cite{martin2024tspm,chen2018scanpath,chen2021predicting,le2016introducing} on model performance and computational costs. Specifically, we compare the Bayesian ConvLSTM used in tSPM-Net~\cite{martin2024tspm}, against the normal ConvLSTM (with the same configuration as the Bayesian ConLSTM) and the CfC variant of the liquid neural network. In all experiments, the VGG19+DeeplabV3 backbone is used, to have a fair comparison with tSPM-Net~\cite{martin2024tspm} that uses VGG19+DeeplabV3 as backbone, while keeping all the training setup unchanged. Table~\ref{tab:rnn comparsion} summarizes the comparison between different recurrent architectures, where the ScanMatch scores for Bayesian ConvLSTM and ConvLSTM are reported as in~\cite{martin2024tspm}. CfC is the most efficient module in terms of computational costs and inference speed, while having better performance compared to the ConvLSTM based modules. Therefore, CfC is chosen as the recurrent module in our model. 
% \begin{table}
% \centering
% \caption{Comparison of different RNN architectures in terms of metric performance and computational cost}
% \label{tab:rnn comparsion}
% \resizebox{\linewidth}{!}{%
% \begin{tabular}{lcccc}
% \toprule
% \textbf{RNN Model} &
%  \textbf{ScanMatch $\uparrow$}
% & \textbf{Time (ms)$\downarrow$}
% & \textbf{\#Params (M)$\downarrow$}
% & \textbf{GFLOPs}$\downarrow$ \\
% \midrule
% Bayesian CONVLSTM  &  0.34 (0.06) & 43.90 & 185.92 & 102.51 \\
% CONVLSTM  & 0.34 (0.06)   & 21.76 & 185.76 &102.51 \\
% CONV-CFC &  \textbf{0.37(0.14)} & \textbf{15.34 }& \textbf{185.71}  & \textbf{99.94} \\
% \bottomrule
% \end{tabular}
% }
% \end{table}

\begin{table}
\vspace{-2ex}
\centering
\caption{RNN models comparison: performance vs. computational cost}
\vspace{-2ex}
\label{tab:rnn comparsion}
\resizebox{\columnwidth}{!}{%
\begin{tabular}{lcccc}
\toprule
\textbf{RNN Model} &
 \textbf{ScanMatch $\uparrow$}
& \textbf{Time (ms)$\downarrow$}
& \textbf{\#Params (M)$\downarrow$}
& \textbf{GFLOPs}$\downarrow$ \\
\midrule
Bayesian CONVLSTM  &  0.34 (0.06) & 43.90 & 185.92 & 102.51 \\
CONVLSTM  & 0.34 (0.06)   & 21.76 & \textbf{185.76} &102.51 \\
% CONV-CFC &  0.37 (0.14) & \textbf{15.34 }& \textbf{185.71}  & 99.94 \\
CFC &  \textbf{0.47 (0.10)} & \textbf{17.43}& 195.41  & \textbf{99.81} \\
\bottomrule
\end{tabular}
}
\vspace{-5ex}
\end{table}

\begin{figure*}[t]
    \centering
    \includegraphics[width=0.99\textwidth]{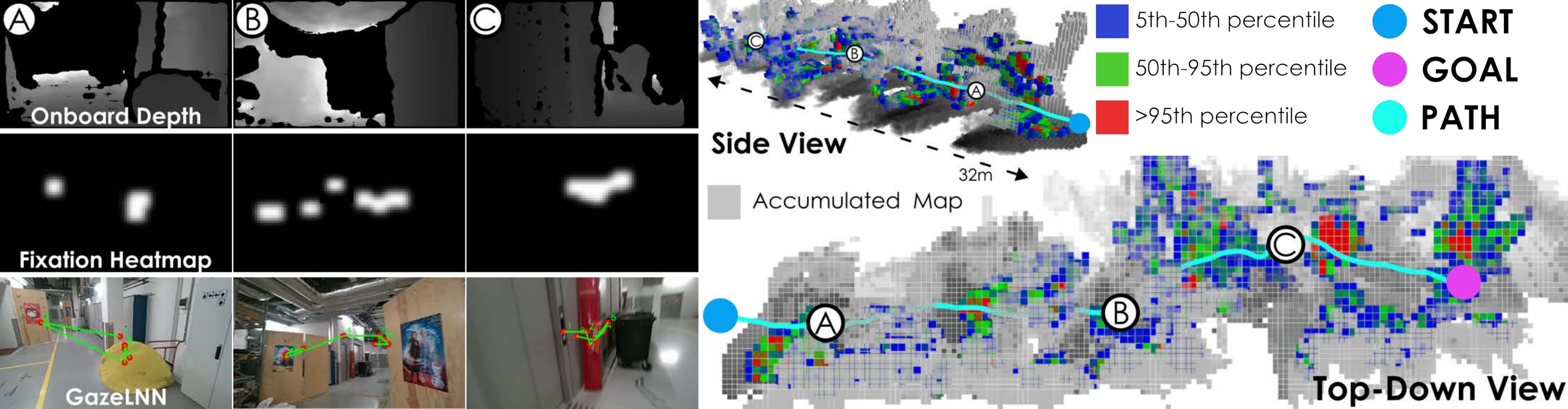}
    \vspace{-2ex}
    \caption{Navigation using the proposed fixation-guided active camera. The camera actively scans the environment (viewpoints A, B, and C), focusing on salient features such as obstacles and visual landmarks. This results in a significantly denser accumulated point cloud and a broader distribution of highly-observed voxels across the operational space.}
    \label{fig:active_camera}
    \vspace{-3ex}
\end{figure*}

\section{EXPERIMENTAL VALIDATION OF FIXATION-GUIDED ROBOT NAVIGATION}
\label{sec:experiments}
To validate the practical applicability of the proposed system beyond fixation prediction, we deploy and evaluate the full pipeline on a physical robotic platform. While the active camera policy was trained entirely within the Aerial Gym simulator, the lightweight nature of GazeLNN and the domain randomization applied during training facilitate transfer to real-world conditions.

The experimental platform is a custom quadrotor, illustrated in \Cref{fig:title_image}, fitted with an IMU and a radar sensor for state estimation. The robot carries an actuated perception module consisting of an Intel RealSense D455 RGB-D camera mounted on a two-axis pan-tilt mechanism at the edge of the frame. Two servo motors provide independent pitch and yaw control, each backed by an integrated potentiometer for closed-loop position feedback, ensuring accurate tracking of the desired camera orientation. The achievable range of motion spans $\pm \SI{45}{\degree}$ in yaw and $\pm \SI{60}{\degree}$ in pitch. The camera delivers synchronized color and depth streams at up to \SI{15}{\Hz}, supporting both scene understanding and spatially-aware obstacle avoidance.

Onboard computation is handled by an NVIDIA Jetson Orin NX 16GB module, which runs the high-level navigation stack, while low-level attitude stabilization and motor control are managed by a PX4 flight controller. In these experiments, GazeLNN runs onboard in real-time, providing live fixation heatmap predictions that guide the camera control policy, which is executed at \SI{10}{\Hz}. All components communicate via ROS as a middleware. We evaluate the system's ability to maintain human fixation-guided behavior while simultaneously performing goal-directed navigation and collision avoidance in unstructured real-world environments.

\begin{figure}[t]
    \centering
    \includegraphics[width=\columnwidth]{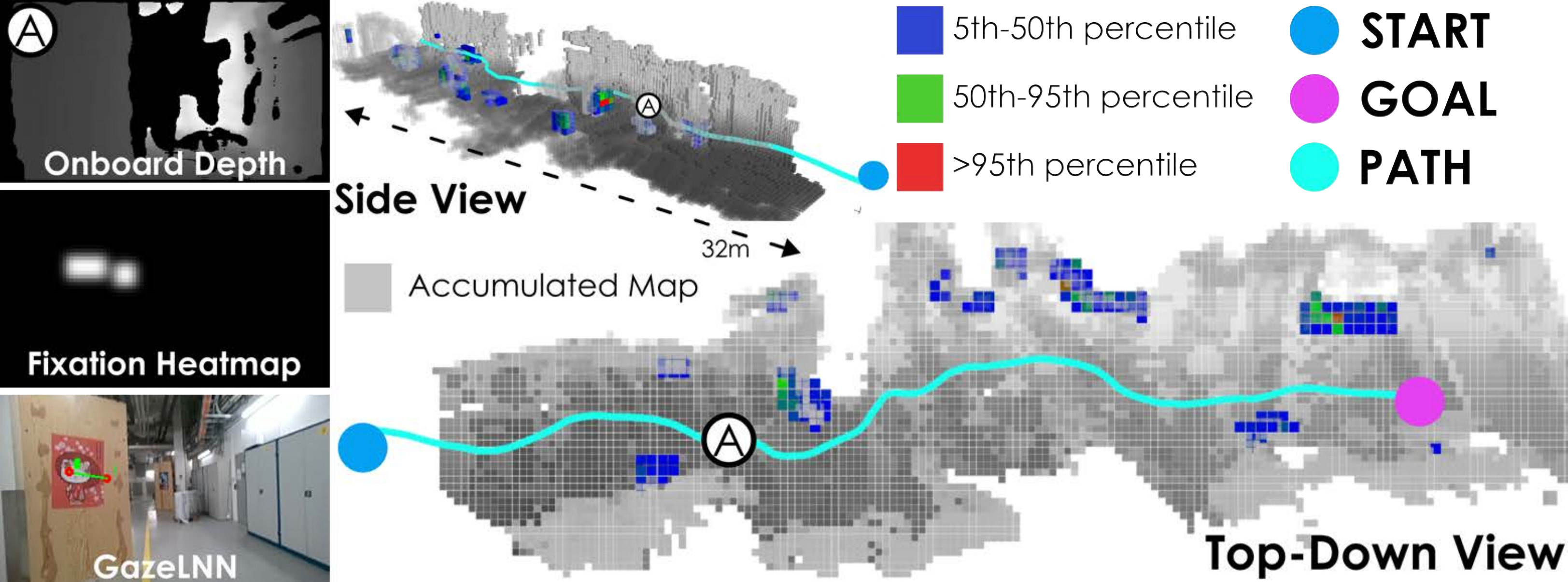}
    \vspace{-5ex}
    \caption{Baseline navigation using a static, forward-facing camera. The left column shows the depth map, generated heatmap, and RGB feed with a centralized fixation point. The right column displays the resulting sparse 3D point cloud accumulation and voxel observation grid. The robot successfully traverses the path, but peripheral scene awareness remains minimal.}
    \label{fig:static_camera}
    \vspace{-4ex}
\end{figure}

\subsection{Qualitative Evaluation of Fixation-guided Behavior}
We compare our proposed active camera policy against a standard static camera baseline in an indoor, unstructured environment that requires continuous maneuvering for collision avoidance, as shown in \Cref{fig:static_camera}. The static camera remains fixed, facing forward. While the robot successfully navigates from the start to the goal, its scene awareness is strictly limited to the narrow corridor directly ahead of it. Because the camera does not actively redirect its field of view, the predicted fixations remain centralized within the camera's frame. Furthermore, the accumulated voxel grid is limited in coverage; significant peripheral regions simply remain completely unmapped due to the restricted viewpoint. Essentially, the baseline captures very little of the surrounding context and does not demonstrate distinct information-gathering behavior, relying solely on forward-facing sensor data to ensure collision-free flight.
% The resulting heatmap is largely static, and the accumulated 3D point cloud is sparse, capturing very little of the surrounding context and broadly not demonstrating a distinct scene information-guided behavior. Essentially, the method relies on depth to ensure collision-free flight.

Conversely, \Cref{fig:active_camera} demonstrates the dynamic behavior of the proposed fixation-guided active vision policy. Instructed by the continuous output of GazeLNN, the pan-tilt mechanism actively directs the camera's field of view toward salient regions, such as visually distinct posters, structural corners, and potential obstacles. The generated scanpaths (shown in green) exhibit human attention-informed movements, sampling multiple points of interest. Consequently, the accumulated voxelized point cloud captures a significantly broader volume of the operational space; rather than increasing local point density, the active system successfully maps peripheral structures and contextual features that the static baseline misses, providing the navigation stack with a vastly richer and more complete spatial representation.
%Consequently, the accumulated point cloud is remarkably denser, providing the navigation stack with a vastly richer representation of the environment.

\subsection{Quantitative Analysis of Scene Exploration}
\begin{table}[t]
    \vspace{-2ex}
    \centering
    \caption{Voxel Observation Statistics for Accumulated Point Clouds}
    \vspace{-2ex}
    \label{tab:voxel_stats}
    \resizebox{\columnwidth}{!}{%
    \begin{tabular}{lccccccc}
        \toprule
        & & \multicolumn{3}{c}{\textbf{Hit Counts}} & \multicolumn{3}{c}{\textbf{Hit Percentiles}} \\
        \cmidrule(lr){3-5} \cmidrule(lr){6-8}
        \textbf{Static Camera} & \begin{tabular}[c]{@{}c@{}}\textbf{Total}\\ \textbf{Voxels}\end{tabular} & \textbf{Min} & \textbf{Max} & \textbf{Avg} & \textbf{5th} & \textbf{50th} & \textbf{95th} \\
        \midrule
        Full Voxel Grid & 37067 & 1 & 311 & 38.7 & 2.0 & 28.0 & 108.0 \\ 
        Fixation Grid   & 873 & 1 & 537 & 65.7 & 2.0 & 34.0 & 267.4 \\
        \midrule
        \textbf{Active Camera} & &  &  &  &  & & \\
        \midrule
        Full Voxel Grid & 55524 & 1 & 357 & 31.4 & 2.0 & 21.0  & 95.0 \\ 
        Fixation Grid   & 6770 & 1 & 756 & 38.7  & 2.0 & 24.0 & 111.0 \\
        \bottomrule
    \end{tabular}%
    }
    \vspace{-5ex}
\end{table}

To quantify the improvement in spatial awareness, we track the observation statistics of the accumulated voxel grid throughout the trajectories. First, as detailed in \Cref{tab:voxel_stats}, the active camera maps significantly more of the environment, accumulating $55,524$ total voxels in the full grid compared to $37,067$ for the static camera. 

More importantly, the active policy dramatically increases task-relevant observation. In the fixation grid, which represents the highly salient areas deemed critical by the GazeLNN network, the active camera policy observes $6,770$ voxels, a nearly eight-fold increase over the static baseline ($873$ voxels). Furthermore, the maximum hit count (defined as the total number of times a specific voxel is observed by the camera sensor) increases from $537$ to $756$, indicating that the active camera policy not only finds more salient features but sustains attention on them longer, providing more robust data for collision avoidance and state estimation.
\section{CONCLUSIONS}
\label{sec: conclusion}
An efficient scanpath prediction model, GazeLNN, is proposed that leverages a liquid neural network as the recurrent module. GazeLNN achieves state-of-the-art performance on the MIT Low Resolution dataset with a Scanmatch score of 0.47, while reducing the computational costs by 99.40\% and having an inference speed-up of $6.42\times$. GazeLNN is further integrated with an active camera control policy trained via \ac{rl} to investigate the role of human attention modeling in robot autonomy. The experiments show that the policy maintains human-fixation-guided perception behavior, pointing the active camera to the fixation locations predicted by GazeLNN while performing a navigation task. Future work shall further investigate the correlation of fixation scanpath, the configuration of the active camera and its motion dynamics, and the reinforcement learning policy. 

\bibliographystyle{IEEEtran} 
\bibliography{reference}

@article{itti2002model,
	title        = {A model of saliency-based visual attention for rapid scene analysis},
	author       = {Itti, Laurent and others},
	year         = 2002,
	journal      = {IEEE Transactions on pattern analysis and machine intelligence},
	publisher    = {Ieee},
	volume       = 20,
	number       = 11,
	pages        = {1254--1259}
}

@incollection{koch1987shifts,
	title        = {Shifts in selective visual attention: towards the underlying neural circuitry},
	author       = {Koch, Christof and others},
	year         = 1987,
	booktitle    = {Matters of intelligence: Conceptual structures in cognitive neuroscience},
	publisher    = {Springer},
	pages        = {115--141}
}

@inproceedings{linardos2021deepgaze,
	title        = {DeepGaze IIE: Calibrated prediction in and out-of-domain for state-of-the-art saliency modeling},
	author       = {Linardos, Akis and others},
	year         = 2021,
	booktitle    = {Proceedings of the IEEE/CVF International Conference on Computer Vision},
	pages        = {12919--12928}
}

@inproceedings{hosseini2025sum,
	title        = {Sum: Saliency unification through mamba for visual attention modeling},
	author       = {Hosseini, Alireza and others},
	year         = 2025,
	booktitle    = {2025 IEEE/CVF Winter Conference on Applications of Computer Vision (WACV)},
	pages        = {1597--1607},
	organization = {IEEE}
}

@article{kummerer2022deepgaze,
	title        = {DeepGaze III: Modeling free-viewing human scanpaths with deep learning},
	author       = {K{\"u}mmerer, Matthias and others},
	year         = 2022,
	journal      = {Journal of Vision},
	publisher    = {The Association for Research in Vision and Ophthalmology},
	volume       = 22,
	number       = 5,
	pages        = {7--7}
}

@inproceedings{chen2018scanpath,
	title        = {Scanpath Prediction for Visual Attention using IOR-ROI LSTM.},
	author       = {Chen, Zhenzhong and others},
	year         = 2018,
	booktitle    = {IJCAI},
	volume       = 2,
	number       = 3,
	pages        = 5
}

@inproceedings{chen2021predicting,
	title        = {Predicting human scanpaths in visual question answering},
	author       = {Chen, Xianyu and others},
	year         = 2021,
	booktitle    = {Proceedings of the IEEE/CVF Conference on Computer Vision and Pattern Recognition},
	pages        = {10876--10885}
}

@article{martin2024tspm,
	title        = {tSPM-Net: A probabilistic spatio-temporal approach for scanpath prediction},
	author       = {Martin, Daniel and others},
	year         = 2024,
	journal      = {Computers \& Graphics},
	publisher    = {Elsevier},
	volume       = 122,
	pages        = 103983
}

@inproceedings{yang2024unifying,
	title        = {Unifying top-down and bottom-up scanpath prediction using transformers},
	author       = {Yang, Zhibo and others},
	year         = 2024,
	booktitle    = {Proceedings of the IEEE/CVF Conference on Computer Vision and Pattern Recognition},
	pages        = {1683--1693}
}

@inproceedings{mondal2023gazeformer,
	title        = {Gazeformer: Scalable, effective and fast prediction of goal-directed human attention},
	author       = {Mondal, Sounak and others},
	year         = 2023,
	booktitle    = {Proceedings of the IEEE/CVF Conference on Computer Vision and Pattern Recognition},
	pages        = {1441--1450}
}

@inproceedings{hasani2021liquid,
	title        = {Liquid time-constant networks},
	author       = {Hasani, Ramin and others},
	year         = 2021,
	booktitle    = {Proceedings of the AAAI Conference on Artificial Intelligence},
	volume       = 35,
	number       = 9,
	pages        = {7657--7666}
}

@article{hasani2022closed,
  title={Closed-form continuous-time neural networks},
  author={Hasani, Ramin and Lechner, Mathias and Amini, Alexander and Liebenwein, Lucas and Ray, Aaron and Tschaikowski, Max and Teschl, Gerald and Rus, Daniela},
  journal={Nature Machine Intelligence},
  volume={4},
  number={11},
  pages={992--1003},
  year={2022},
  publisher={Nature Publishing Group UK London}
}

@article{stewart2020review,
	title        = {A review of interactions between peripheral and foveal vision},
	author       = {Stewart, Emma EM and others},
	year         = 2020,
	journal      = {Journal of vision},
	publisher    = {The Association for Research in Vision and Ophthalmology},
	volume       = 20,
	number       = 12,
	pages        = {2--2}
}

@inproceedings{mohammed2025unified,
	title        = {Unified attention modeling for efficient free-viewing and visual search via shared representations},
	author       = {Mohammed, Fatma Youssef and others},
	year         = 2025,
	booktitle    = {2025 IEEE International Conference on Development and Learning (ICDL)},
	pages        = {1--8},
	organization = {IEEE}
}

@article{liu2018intriguing,
	title        = {An intriguing failing of convolutional neural networks and the coordconv solution},
	author       = {Liu, Rosanne and others},
	year         = 2018,
	journal      = {Advances in neural information processing systems},
	volume       = 31
}

@article{sun2019visual,
	title        = {Visual scanpath prediction using IOR-ROI recurrent mixture density network},
	author       = {Sun, Wanjie and others},
	year         = 2019,
	journal      = {IEEE transactions on pattern analysis and machine intelligence},
	publisher    = {IEEE},
	volume       = 43,
	number       = 6,
	pages        = {2101--2118}
}

@article{xu2014predicting,
	title        = {Predicting human gaze beyond pixels},
	author       = {Xu, Juan and others},
	year         = 2014,
	journal      = {Journal of vision},
	publisher    = {The Association for Research in Vision and Ophthalmology},
	volume       = 14,
	number       = 1,
	pages        = {28--28}
}

@article{judd2011fixations,
	title        = {Fixations on low-resolution images},
	author       = {Judd, Tilke and others},
	year         = 2011,
	journal      = {Journal of Vision},
	publisher    = {The Association for Research in Vision and Ophthalmology},
	volume       = 11,
	number       = 4,
	pages        = {14--14}
}

@article{fahimi2021metrics,
	title        = {On metrics for measuring scanpath similarity},
	author       = {Fahimi, Ramin and others},
	year         = 2021,
	journal      = {Behavior Research Methods},
	publisher    = {Springer},
	volume       = 53,
	number       = 2,
	pages        = {609--628}
}

@article{malczyk2026reinforcement,
	title        = {Reinforcement Learning for Active Perception in Autonomous Navigation},
	author       = {Malczyk, Grzegorz and others},
	year         = 2026,
	journal      = {arXiv preprint arXiv:2602.01266}
}

@article{kulkarni2025aerial,
	title        = {Aerial gym simulator: A framework for highly parallelized simulation of aerial robots},
	author       = {Kulkarni, Mihir and others},
	year         = 2025,
	journal      = {IEEE Robotics and Automation Letters},
	publisher    = {IEEE}
}

@inproceedings{petrenko2020sample,
	title        = {Sample factory: Egocentric 3d control from pixels at 100000 fps with asynchronous reinforcement learning},
	author       = {Petrenko, Aleksei and others},
	year         = 2020,
	booktitle    = {International Conference on Machine Learning},
	pages        = {7652--7662},
	organization = {PMLR}
}

@inproceedings{espeholt2018impala,
	title        = {Impala: Scalable distributed deep-rl with importance weighted actor-learner architectures},
	author       = {Espeholt, Lasse and others},
	year         = 2018,
	booktitle    = {International conference on machine learning},
	pages        = {1407--1416},
	organization = {PMLR}
}

@article{le2016introducing,
	title        = {Introducing context-dependent and spatially-variant viewing biases in saccadic models},
	author       = {Le Meur, Olivier and others},
	year         = 2016,
	journal      = {Vision research},
	publisher    = {Elsevier},
	volume       = 121,
	pages        = {72--84}
}

@article{simeoni2025dinov3,
	title        = {Dinov3},
	author       = {Sim{\'e}oni, Oriane and others},
	year         = 2025,
	journal      = {arXiv preprint arXiv:2508.10104}
}

@inproceedings{howard2019searching,
	title        = {Searching for mobilenetv3},
	author       = {Howard, Andrew and others},
	year         = 2019,
	booktitle    = {Proceedings of the IEEE/CVF international conference on computer vision},
	pages        = {1314--1324}
}

@inproceedings{yurtkulu2019semantic,
	title        = {Semantic segmentation with extended DeepLabv3 architecture},
	author       = {Yurtkulu, Salih Can and others},
	year         = 2019,
	booktitle    = {2019 27th signal processing and communications applications conference (SIU)},
	pages        = {1--4},
	organization = {IEEE}
}

@inproceedings{he2016deep,
	title        = {Deep residual learning for image recognition},
	author       = {He, Kaiming and others},
	year         = 2016,
	booktitle    = {Proceedings of the IEEE conference on computer vision and pattern recognition},
	pages        = {770--778}
}

@article{simonyan2014very,
	title        = {Very deep convolutional networks for large-scale image recognition},
	author       = {Simonyan, Karen and others},
	year         = 2014,
	journal      = {arXiv preprint arXiv:1409.1556}
}

@inproceedings{kulkarni2023task,
	title        = {Task-driven compression for collision encoding based on depth images},
	author       = {Kulkarni, Mihir and others},
	year         = 2023,
	booktitle    = {International Symposium on Visual Computing},
	pages        = {259--273},
	organization = {Springer}
}

@article{malczyk2025semantically,
	title        = {Semantically-Driven Deep Reinforcement Learning for Inspection Path Planning},
	author       = {Malczyk, Grzegorz and others},
	year         = 2025,
	journal      = {IEEE Robotics and Automation Letters},
	publisher    = {IEEE}
}

@inproceedings{chang2010mobile,
	title        = {Mobile robot vision navigation \& localization using gist and saliency},
	author       = {Chang, Chin-Kai and others},
	year         = 2010,
	booktitle    = {2010 IEEE/RSJ International Conference on Intelligent Robots and Systems},
	pages        = {4147--4154},
	organization = {IEEE}
}

@article{guo2021motion,
	title        = {Motion saliency-based collision avoidance for mobile robots in dynamic environments},
	author       = {Guo, Binghua and others},
	year         = 2021,
	journal      = {IEEE Transactions on Industrial Electronics},
	publisher    = {IEEE},
	volume       = 69,
	number       = 12,
	pages        = {13203--13212}
}

@article{rasouli2020attention,
	title        = {Attention-based active visual search for mobile robots},
	author       = {Rasouli, Amir and others},
	year         = 2020,
	journal      = {Autonomous Robots},
	publisher    = {Springer},
	volume       = 44,
	number       = 2,
	pages        = {131--146}
}

@article{potapova2017survey,
	title        = {Survey of recent advances in 3D visual attention for robotics},
	author       = {Potapova, Ekaterina and others},
	year         = 2017,
	journal      = {The International Journal of Robotics Research},
	publisher    = {SAGE Publications Sage UK: London, England},
	volume       = 36,
	number       = 11,
	pages        = {1159--1176}
}

@article{bajcsy2018revisiting,
	title        = {Revisiting active perception},
	author       = {Bajcsy, Ruzena and others},
	year         = 2018,
	journal      = {Autonomous Robots},
	publisher    = {Springer},
	volume       = 42,
	number       = 2,
	pages        = {177--196}
}

@inproceedings{dang2018visual,
	title        = {Visual saliency-aware receding horizon autonomous exploration with application to aerial robotics},
	author       = {Dang, Tung and others},
	year         = 2018,
	booktitle    = {2018 IEEE international conference on robotics and automation (ICRA)},
	pages        = {2526--2533},
	organization = {IEEE}
}

@inproceedings{liang2024visarl,
	title        = {Visarl: Visual reinforcement learning guided by human saliency},
	author       = {Liang, Anthony and others},
	year         = 2024,
	booktitle    = {2024 IEEE/RSJ International Conference on Intelligent Robots and Systems (IROS)},
	pages        = {2907--2912},
	organization = {IEEE}
}

@article{ma2018saliency,
	title        = {A saliency-based reinforcement learning approach for a UAV to avoid flying obstacles},
	author       = {Ma, Zhaowei and others},
	year         = 2018,
	journal      = {Robotics and Autonomous Systems},
	publisher    = {Elsevier},
	volume       = 100,
	pages        = {108--118}
}

@article{pfeiffer2022visual,
	title        = {Visual attention prediction improves performance of autonomous drone racing agents},
	author       = {Pfeiffer, Christian and others},
	year         = 2022,
	journal      = {Plos one},
	publisher    = {Public Library of Science San Francisco, CA USA},
	volume       = 17,
	number       = 3,
	pages        = {e0264471}
}

@article{frintrop2008attentional,
	title        = {Attentional landmarks and active gaze control for visual SLAM},
	author       = {Frintrop, Simone and others},
	year         = 2008,
	journal      = {IEEE Transactions on Robotics},
	publisher    = {IEEE},
	volume       = 24,
	number       = 5,
	pages        = {1054--1065}
}

@article{rasouli2017integrating,
	title        = {Integrating three mechanisms of visual attention for active visual search},
	author       = {Rasouli, Amir and others},
	year         = 2017,
	journal      = {arXiv preprint arXiv:1702.04292}
}

@incollection{lecun2002efficient,
  title={Efficient backprop},
  author={LeCun, Yann  and others},
  booktitle={Neural networks: Tricks of the trade},
  pages={9--50},
  year={2002},
  publisher={Springer}
}
\end{document}